\documentclass[10pt,journal,compsoc]{IEEEtran}
\usepackage{amsmath,amsfonts}
\usepackage{algorithmic}
\usepackage{array}
\usepackage[caption=false,font=normalsize,labelfont=sf,textfont=sf]{subfig}
\usepackage{textcomp}
\usepackage{stfloats}
\usepackage{url}
\usepackage{verbatim}
\usepackage{graphicx}
\usepackage{graphicx} 
\usepackage{amsmath}
\usepackage{amsfonts}
\usepackage{multirow}
\usepackage{color,xcolor}
\usepackage{bbding}
\usepackage{graphicx}
\usepackage{amsmath}
\usepackage{amssymb}
\usepackage{booktabs}

\usepackage{enumitem}
\usepackage{multirow}

\usepackage{algorithm}
\usepackage{algorithmic}

\usepackage{newfloat}
\usepackage{listings}
\DeclareCaptionStyle{ruled}{labelfont=normalfont,labelsep=colon,strut=off} % DO NOT CHANGE THIS
\floatstyle{ruled}
\newfloat{listing}{tb}{lst}{}
\floatname{listing}{Listing}

\usepackage[colorlinks,
            linkcolor=red,       %%修改此处为你想要的颜色
            anchorcolor=blue,  %%修改此处为你想要的颜色
            citecolor=blue,        %%修改此处为你想要的颜色，例如修改blue为red
            ]{hyperref}

\usepackage{amsthm} % 如果要使用proof语句就必须要引入这个包

\usepackage{orcidlink}

\ifCLASSOPTIONcompsoc
  \usepackage[nocompress]{cite}
\else
  \usepackage{cite}
\fi
\usepackage{enumitem}
\begin{document}

\title{Predicting and Enhancing the Fairness of DNNs with the Curvature of Perceptual Manifolds}

\author{Yanbiao Ma~\orcidlink{0000-0002-8472-1475},
        Licheng Jiao~\orcidlink{0000-0003-3354-9617},~\IEEEmembership{Fellow,~IEEE,}
        Fang Liu~\orcidlink{0000-0002-5669-9354},~\IEEEmembership{Senior Member,~IEEE,}
        Maoji Wen,
        Lingling Li~\orcidlink{0000-0002-6130-2518},~\IEEEmembership{Senior Member,~IEEE,}
        Wenping Ma~\orcidlink{0000-0001-8872-2195},~\IEEEmembership{Senior Member,~IEEE,}
        Shuyuan Yang~\orcidlink{0000-0002-4796-5737},~\IEEEmembership{Senior Member,~IEEE,}
        Xu Liu~\orcidlink{0000-0002-8780-5455},~\IEEEmembership{Senior Member,~IEEE,}
        Puhua Chen~\orcidlink{0000-0001-5472-1426},~\IEEEmembership{Senior Member,~IEEE}
        %}% <-this % stops a space

\thanks{This work was supported in part by the Key Scientific Technological Innovation Research Project of the Ministry of Education, the Joint Funds of the National Natural Science Foundation of China (U22B2054), the National Natural Science Foundation of China (62076192, 61902298, 61573267, 61906150, and 62276199), the 111 Project, the Program for Cheung Kong Scholars and Innovative Research Team in University (IRT 15R53), the Science and Technology Innovation Project from the Chinese Ministry of Education, the Key Research and Development Program in Shaanxi Province of China (2019ZDLGY03-06), and the China Postdoctoral Fund (2022T150506). 

\emph{(Corresponding author: Licheng Jiao.)}}% <-this % stops a space
\thanks{The authors are with the Key Laboratory of Intelligent Perception and Image Understanding of the Ministry of Education of China, International Research Center of Intelligent Perception and Computation, School of Artificial Intelligence, Xidian University, Xian 710071, China (e-mail: ybmamail@stu.xidian.edu.cn; lchjiao@mail.xidian.edu.cn).}% <-this % stops a space
%\fi
}

\iffalse
\IEEEcompsocitemizethanks{\IEEEcompsocthanksitem M. Shell was with the Department
of Electrical and Computer Engineering, Georgia Institute of Technology, Atlanta,
GA, 30332.\protect\\
E-mail: see http://www.michaelshell.org/contact.html
\IEEEcompsocthanksitem J. Doe and J. Doe are with Anonymous University.}% <-this % stops a space
\thanks{Manuscript received April 19, 2005; revised August 26, 2015.}}
\fi

\iffalse
\markboth{Journal of \LaTeX\ Class Files,~Vol.~14, No.~8, August~2015}%
{Shell \MakeLowercase{\textit{et al.}}: Bare Advanced Demo of IEEEtran.cls for IEEE Computer Society Journals}
\fi

\IEEEtitleabstractindextext{%
\begin{abstract}
To address the challenges of long-tailed classification, researchers have proposed several approaches to reduce model bias, most of which assume that classes with few samples are weak classes. However, recent studies have shown that tail classes are not always hard to learn, and model bias has been observed on sample-balanced datasets, suggesting the existence of other factors that affect model bias. In this work, we first establish a geometric perspective for analyzing model fairness and then systematically propose a series of geometric measurements for perceptual manifolds in deep neural networks. Subsequently, we comprehensively explore the effect of the geometric characteristics of perceptual manifolds on classification difficulty and how learning shapes the geometric characteristics of perceptual manifolds. An unanticipated finding is that the correlation between the class accuracy and the separation degree of perceptual manifolds gradually decreases during training, while the negative correlation with the curvature gradually increases, implying that curvature imbalance leads to model bias. We thoroughly validate this finding across multiple networks and datasets, providing a solid experimental foundation for future research. We also investigate the convergence consistency between the loss function and curvature imbalance, demonstrating the lack of curvature constraints in existing optimization objectives. Building upon these observations, we propose curvature regularization to facilitate the model to learn curvature-balanced and flatter perceptual manifolds. Evaluations on multiple long-tailed and non-long-tailed datasets show the excellent performance and exciting generality of our approach, especially in achieving significant performance improvements based on current state-of-the-art techniques. Our work opens up a geometric analysis perspective on model bias and reminds researchers to pay attention to model bias on non-long-tailed and even sample-balanced datasets.
\end{abstract}

}

\maketitle

\IEEEdisplaynontitleabstractindextext

\IEEEpeerreviewmaketitle

\ifCLASSOPTIONcompsoc
\IEEEraisesectionheading{\section{Introduction}\label{sec:introduction}}
\else
\section{Introduction}
\label{sec:introduction}
\fi

\IEEEPARstart{T}{he} imbalance of sample numbers in the dataset gives rise to the challenge of long-tailed visual recognition. Most previous works assume that head classes are always easier to be learned than tail classes, e.g., class re-balancing \cite{paper4,paper2,paper14,paper3,paper36,paper37,paper9}, information augmentation \cite{paper38,paper39,paper40,paper41,paper42,paper12,paper43,paper44,paper47}, decoupled training \cite{paper35,paper29,paper1,paper10,paper11,paper46}, and ensemble learning \cite{paper5,paper30,paper31,paper32,paper33,paper7,paper34} have been proposed to improve the performance of tail classes. However, recent studies \cite{paper27, paper28} have shown that classification difficulty is not always correlated with the number of samples, e.g., the performance of some tail classes is even higher than that of the head classes. Also, \cite{paper45} observes differences in model performance across classes on non-long-tailed data, and even on balanced data. Therefore, it is necessary to explore the impact of other inherent characteristics of the data on the classification difficulty, and then improve the overall performance by mitigating the model bias under multiple sample number distribution scenarios.

\begin{figure}[t]
%\vskip -0.15in
\centering
\centerline{\includegraphics[width=1\columnwidth]{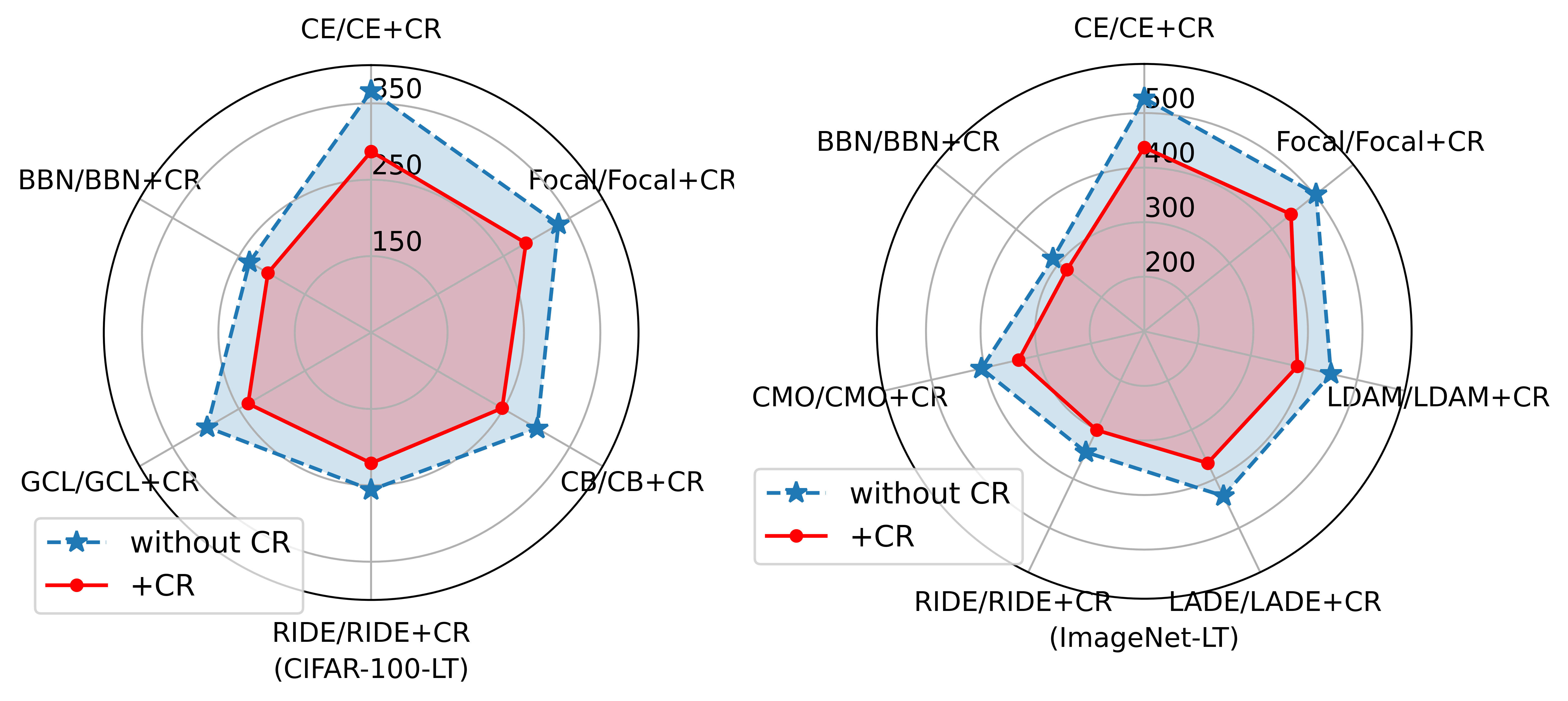}}
\vskip -0.1in
\caption{Curvature regularization reduces the model bias present in multiple methods on CIFAR-100-LT and ImageNet-LT. The model bias is measured with the variance of the accuracy of all classes, and it is zero when the accuracy of each class is the same.}
\label{fig1}
%\vskip -0.2in
\end{figure}

\iffalse
\renewcommand{\thefootnote}{}
\footnotetext{This work was supported in part by
the Key Scientific Technological Innovation Research Project by Ministry of Education,
the State Key Program and the Foundation for Innovative Research Groups of the National Natural Science Foundation of China (61836009),
the Major Research Plan of the National Natural Science Foundation of China (91438201, 91438103, and 91838303),
the National Natural Science Foundation of China (U22B2054, U1701267, 62076192, 62006177, 61902298, 61573267, 61906150, and 62276199),
the 111 Project,
the Program for Cheung Kong Scholars and Innovative Research Team in University (IRT 15R53),
the ST Innovation Project from the Chinese Ministry of Education,
the Key Research and Development Program in Shaanxi Province of China(2019ZDLGY03-06),
the National Science Basic Research Plan in Shaanxi Province of China(2022JQ-607),
the China Postdoctoral fund(2022T150506),
the Scientific Research Project of Education Department In Shaanxi Province of China (No.20JY023),
the National Natural Science Foundation of China (No. 61977052).}
\fi

Focal loss \cite{paper3} utilizes the DNN's prediction confidence on instances to evaluate the instance-level difficulty. \cite{paper27} argues that for long-tailed problems, determining class-level difficulty is more important than determining instance-level difficulty, and therefore defines classification difficulty by evaluating the accuracy of each class in real-time. However, both methods rely on the model output and still cannot explain why the model performs well in some classes and poorly in others. Similar to the number of samples, we would like to propose a measure that relies solely on the data itself to model class-level difficulty, which helps to understand how deep neural networks learn from the data. The effective number of samples \cite{paper4} tries to characterize the diversity of features in each class, but it introduces hyperparameters and would not work in balanced dataset.

\begin{figure*}[t]
%\vskip -0.15in
\centering
\centerline{\includegraphics[width=2\columnwidth]{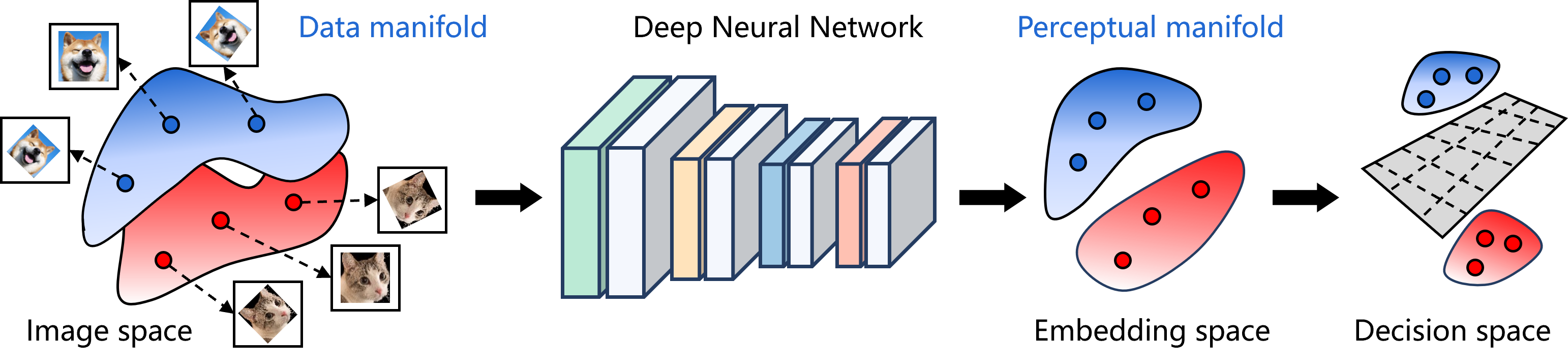}}
\vskip -0.1in
\caption{The geometric perspective of data classification involves each class of data distributed around a submanifold. In image space, multiple submanifolds may be intertwined. Deep neural networks untangle these submanifolds and separate them from each other through layer-wise mappings, facilitating classification. The class perceptual manifolds in the embedding space are mapped into the decision space for classification, so the geometric complexity of the perceptual manifolds may directly affect the classification performance.}
\label{fig100}
%\vskip -0.2in
\end{figure*}

Natural images usually obey the manifold distribution law \cite{paper48,paper49}, i.e., samples of each class are distributed near a low-dimensional manifold in the high-dimensional space. The manifold consisting of features in the embedding space is called a perceptual manifold \cite{paper50}. As shown in Fig.\ref{fig100}, the classification task is equivalent to distinguishing each perceptual manifold, which has a series of geometric characteristics. A well-trained deep neural network achieves classification by untangling the perceptual manifolds and separating them. We speculate that some geometric characteristics may affect the classification difficulty, and therefore conduct an in-depth study. 

\textbf{The main contributions of our work are:} %\textbf{(1)} We systematically propose a series of measurements for the geometric characteristics of point cloud perceptual manifolds in deep neural networks (Sec \ref{sec3}). \textbf{(2)} The effect of learning on the separation degree (Sec \ref{sec4.1}) and curvature (Sec \ref{sec4.2}) of perceptual manifolds is explored. We find that the correlation between separation degree and class accuracy decreases with training, while the negative correlation between curvature and class accuracy increases with training (Sec \ref{sec4.3}), implying that existing methods can only mitigate the effect of separation degree among perceptual manifolds on model bias, while ignoring the effect of perceptual manifold complexity on model bias. \textbf{(3)} Curvature regularization is proposed to facilitate the model to learn curvature-balanced and flatter feature manifolds, thus improving the overall performance (Sec \ref{sec5}). Our approach effectively reduces the model bias on multiple long-tailed (Fig \ref{fig1}) and non-long-tailed datasets (Fig \ref{fig8}), showing excellent performance (Sec \ref{sec6}).

\begin{itemize}[]
\item[(1)] We introduce a novel geometric perspective to assess the fairness of models. Under this perspective, we systematically propose a series of metrics for measuring the geometric characteristics of point cloud perceptual manifolds in deep neural networks, including the volume, separability, and curvature of perceptual manifolds  (Sec \ref{sec3}). These metrics provide tools for quantitative analysis.

\item[(2)] We found that even on balanced datasets, there exists a significant negative correlation between the curvature of the perceptual manifold corresponding to each class and the model's class accuracy (Sec \ref{newsec4}). This discovery provides a new mechanism for explaining and evaluating the fairness of the model.

\item[(3)] We comprehensively investigated the dynamics of the geometric characteristics of perceptual manifolds. In particular, we explored the effects of learning on the separability (Sec \ref{sec4.1}) and curvature (Sec \ref{sec4.2}) of perceptual manifolds. We find that the correlation between separation degree and class accuracy decreases with training, while the negative correlation between curvature and class accuracy increases with training (Sec \ref{sec4.3}), implying that existing methods can only mitigate the effect of separation degree among perceptual manifolds on model bias, while ignoring the effect of perceptual manifold complexity on model bias.

\item[(4)] Curvature regularization is proposed to facilitate the model to learn curvature-balanced and flatter perceptual manifolds, thus mitigating model bias (Fig.\ref{fig1} and Fig.\ref{fig8}) while improving its overall performance (Sec \ref{sec5}). Our approach shows excellent performance on multiple long-tailed and non-long-tailed datasets (Sec \ref{sec6}).
\end{itemize}

This work is an extension of the CVPR 2023 paper. Compared to the initial version, Section \ref{sec3} has been expanded to include proofs of the properties of the proposed perceptual manifold separability. More simulation examples are provided to elucidate the measure of perceptual manifold separability that we propose. We added Section \ref{newsec4}, where we comprehensively unveil the correlation between the curvature of perceptual manifolds generated by different layers of deep neural networks and model fairness. The experimental results of $13$ models across three datasets strongly validate our viewpoint that the curvature of perceptual manifolds can predict model fairness. This provides a solid experimental foundation for improving model fairness. We rewrote Section \ref{sec4}, adding more experimental results and analysis. More importantly, we included an exploration of the convergence consistency between the loss function and curvature, as well as curvature imbalance, revealing the lack of curvature constraints in existing optimization objectives. In Section \ref{sec6}, we added detailed steps for deriving curvature regularization and enhanced the motivation. In the experimental section, we added a subsection to introduce the role of curvature regularization in reducing model bias and curvature imbalance. Code published at: \url{https://github.com/mayanbiao1234/Geometric-metrics-for-perceptual-manifolds}.

%%%%%%%%%%%%%%%%%%%%%%%%%%%%%%%%%%%%%%%%%%%%%%%%%%%%%%
%%%%%%%%%%%%%%%%  第二章 %%%%%%%%%%%%%%%%%%%%%%%%%%%%%%%%%%
%%%%%%%%%%%%%%%%%%%%%%%%%%%%%%%%%%%%%%%%%%%%%%%%%%%%%%

\section{Related Work}
%We present a comprehensive survey of recent advances in long-tailed visual classification in Appendix \textcolor{red}{A}.
In practice, the dataset usually tends to follow a long-tailed distribution, which leads to models with very large variances in performance on each class. It should be noted that most researchers default to the main motivation for long-tail visual recognition is that classes with few samples are always weak classes. Therefore, numerous methods have been proposed to improve the performance of the model on tail classes. \cite{paper58} divides these methods into three fields, namely class rebalancing \cite{paper37,paper45,paper36,paper27,paper4,paper3,paper59,paper60,paper61,paper62,paper63, paper31,paper64,paper65,paper1}, information augmentation \cite{paperma,paperma1,paper38,paper39,paper42,paper10,paper41,paper8,paper72,paper73,paper74,paper71,paper43}, and module improvement \cite{paper75,paper76,paper46,paper13,paper78,paper5,paper31,paper77,paper7,paper34}. Unlike the above, \cite{paper27} and \cite{paper28} observe that the number of samples in the class does not exactly show a positive correlation with the accuracy, and the accuracy of some tail classes is even higher than the accuracy of the head class. Therefore, they propose to use other measures to gauge the learning difficulty of the classes rather than relying on the sample number alone. In the following, we first present past research up to \cite{paper27}\cite{paper28} and lead to our work.

\subsection*{Class-Difficulty Based Methods}
\label{sec2.4}

\textbf{The study of class difficulty is most relevant to our work}. The methods in the three domains presented above almost all assume that classes with few samples are the most difficult classes to be learned, and therefore more attention is given to these classes. However, recent studies \cite{paper27,paper28} have observed that the performance of some tail classes is even higher than that of the head classes, and that the performance of different classes varies on datasets with perfectly balanced samples. These phenomena suggest that the sample number is not the only factor that affects the performance of classes. The imbalance in class performance is referred to as the ``bias'' of the model, and \cite{paper27} defines the model bias as $$bias=\max(\frac{\max_{c=1}^NA_c}{\min_{c'=1}^NA_{c'}+\varepsilon}-1,0),$$ where $A_c$ denotes the accuracy of the $c$-th class. When the accuracy of each class is identical, bias = 0. \cite{paper27} computes the difficulty of class c using $1-A_c$ and calculates the weights of the loss function using a nonlinear function of class difficulty. Unlike \cite{paper27}, \cite{paper28} proposes a model-independent measure of classification difficulty, which directly utilizes the data matrix to calculate the semantic scale of each class to represent the classification difficulty. As with the sample number, model-independent measures can help us understand how deep neural networks learn from data. When we get data from any domain, if we can measure the difficulty of each class directly from the data, we can guide the researchers to collect the difficult classes in a targeted manner instead of blindly, greatly facilitating the efficiency of applying AI in practice. 

In this work, we propose to consider the classification task as the classification of perceptual manifolds. The influence of the geometric characteristics of the perceptual manifold on the classification difficulty is further analyzed, and feature learning with curvature balanced is proposed.

\section{The Geometry of Perceptual Manifold}
\label{sec3}
In this section, we systematically propose a series of geometric measures for perceptual manifolds in deep neural networks, and conduct simulation tests and analyses.

\subsection{Perceptual Manifold}
\label{sec3.1}
A perceptual manifold is generated when neurons are stimulated by objects with different physical characteristics from the same class. Sampling along the different dimensions of the manifold corresponds to changes in specific physical characteristics. It has been shown \cite{paper48,paper49} that the features extracted by deep neural networks obey the manifold distribution law. That is, features from the same class are distributed near a low-dimensional manifold in the high-dimensional feature space. Given data $X=[x_1,\dots,x_m]$ from the same class and a deep neural network $M\!odel=\{f(x,\theta_1),g(z,\theta_2) \}$, where $f(x,\theta_1)$ represents a feature sub-network with parameters $\theta_1$ and $g(z,\theta_2)$ represents a classifier with parameters $\theta_2$. Extract the p-dimensional features $Z=[z_1,\dots,z_m]\in \mathbb{R}^{p\times m}$ of $X$ with the trained model, where $z_i=f(x_i,\theta_1)\in \mathbb{R}^p$. Assuming that the features $Z$ belong to class $c$, the $m$ features form a $p$-dimensional point cloud manifold $M^c$, which is called a \textbf{class perceptual manifold} \cite{paper51}. 

\begin{algorithm*}[t]
\caption{Pseudocode for The Volume of Perceptual Manifold}
\label{alg2}
%\footnotesize{
\textbf{Input:} Training set $D = \left\{ {\left( {{x_i},{y_i}} \right)} \right\}_{i = 1}^M$ with the total number $C$ of classes. A CNN $\{f(x,\theta_1),g(z,\theta_2)\}$, where $f(\cdot)$ and $g(\cdot)$ denote the feature sub-network and classifier, respectively. \\
\textbf{Output:} The volume of all perceptual manifolds.
\begin{algorithmic}[1] %[1] enables line numbers
\FOR{$j=1$ to $C$}
   \STATE Select the sample set ${D_j} = \left\{ {\left( {{x_i},{y_i}} \right)} \right\}_{i = 1}^{{m_j}}$ for class $j$ from $D$, ${m_j}$ is the number of samples for class $j$.
   \STATE Calculate the feature embedding $Z_j=\{z_i \mid z_i = f(x_i,\theta_1)\}_{i=1}^{m_j}$ of $D_j$, ${Z_j} = \left[ {{z_1},{z_2}, \ldots ,{z_{{m_j}}}} \right] \in {\mathbb{R}^{p \times {m_j}}}$.
   \STATE ${Z_j} = {Z_j} - N\!um\!P\!y.mean\left( {{Z_j}, {\rm{1}}} \right)$.
   \STATE Calculate the covariance matrix ${\Sigma _j} = \frac{1}{{{m_j}}}{Z_j}Z_j^T$.
   \STATE Calculate the volume $V\!ol\left( {{\Sigma _j}} \right) = \frac{1}{2}{\log _2}\det \left( {I + {\Sigma _j}} \right)$ of the perceptual manifold corresponding to class $j$.
   \ENDFOR
\end{algorithmic}
\end{algorithm*}

\subsection{The Volume of Perceptual Manifold}
\label{sec3.2}
We measure the volume of the perceptual manifold $M^c$ by calculating the size of the subspace spanned by the features $z_1,\dots,z_m$. First, the sample covariance matrix of $Z$ can be estimated as $\Sigma _{Z}=\mathbb{E}[\frac{1}{n} \sum_{i=1}^{n}z_iz_i^T]=\frac{1}{n}Z\!Z^T  \in \mathbb{R}^{p\times p}.$
Diagonalize the covariance matrix $\Sigma _{Z}$ as $U\!DU^T$, where $D=diag(\lambda_1,\dots,\lambda_p)$ and $U=[u_1,\dots,u_p]\in \mathbb{R}^{p\times p}$. $\lambda_i$ and $u_i$ denote the $i$-th eigenvalue of $\Sigma _{Z}$ and its corresponding eigenvector, respectively. Let the singular value of matrix $Z$ be $\sigma_i=\sqrt{\lambda_i} (i=1,\dots,p)$. According to the geometric meaning of singular value \cite{paper52}, the volume of the space spanned by vectors $z_1,\dots,z_m$ is proportional to the product of the singular values of matrix $Z$, i.e., $V\!ol(Z)\propto  {\textstyle \prod_{i=1}^{p}}\sigma_i=\sqrt{ {\textstyle \prod_{i=1}^{p}}\lambda_i}$. Considering $\lambda_1\lambda_2\cdots \lambda_p=\det(\Sigma_Z)$, the volume of the perceptual manifold is therefore denoted as $V\!ol(Z)\propto \sqrt{\det(\frac{1}{m}Z\!Z^T)}$.

However, when $\frac{1}{m}Z\!Z^T$ is a non-full rank matrix, its determinant is $0$. For example, the determinant of a planar point set located in three-dimensional space is 0 because its covariance matrix has zero eigenvalues, but obviously the volume of the subspace tensed by the point set in the plane is non-zero. We want to obtain the ``area'' of the planar point set, which is a generalized volume. We avoid the non-full rank case by adding the unit matrix $I$ to the covariance matrix $\frac{1}{m}Z\!Z^T$. $I+\frac{1}{m}Z\!Z^T$ is a positive definite matrix with eigenvalues $\lambda_i+1 (i=1,\dots,p)$. The above operation enables us to calculate the volume of a low-dimensional manifold embedded in high-dimensional space. The volume $V\!ol(Z)$ of the perceptual manifold is proportional to $\sqrt{\det(I+\frac{1}{m}Z\!Z^T)}$. Considering the numerical stability, we further perform a logarithmic transformation on $\sqrt{\det(I+\frac{1}{m}Z\!Z^T)}$ and define the volume of the perceptual manifold as $$V\!ol(Z)=\frac{1}{2}\log_2 \det(I+\frac{1}{m}(Z-Z_{mean})(Z-Z_{mean})^T),$$ where $Z_{mean}$ is the mean of $Z$. When $m>1$, $V\!ol(Z>0$. Since $I+\frac{1}{m}(Z-Z_{mean})(Z-Z_{mean})^T$ is a positive definite matrix, its determinant is greater than 0. In the following, the degree of separation between perceptual manifolds will be proposed based on the volume of perceptual manifolds.

\subsection{The Separation Degree of Perceptual Manifold}
\label{sec3.3}
%\iffalse
Euclidean or cosine distances between class centers are often applied to measure inter-class distances, and these two distances are also commonly used as loss functions when constructing sample pairs. However, maximizing the distance between proxy points or samples cannot keep a class away from all the remaining classes at the same time, and the distance between class centers does not reflect the degree of overlap of the distribution. In this section, we propose a measure of the separation degree between perceptual manifolds.
%\fi

Given the perceptual manifolds $M^1$ and $M^2$, they consist of point sets $Z_1=[z_{1,1},\dots,z_{1,m_1}]\in \mathbb{R}^{p\times m_1}$ and $Z_2=[z_{2,1},\dots,z_{2,m_2}]\in \mathbb{R}^{p\times m_2}$, respectively. The volumes of $M^1$ and $M^2$ are calculated as $V\!ol(Z_1)$ and $V\!ol(Z_2)$. Consider the following case, assuming that $M^1$ and $M^2$ have partially overlapped, when $V\!ol(Z_1)\ll V\!ol(Z_2)$, it is obvious that the overlapped volume accounts for a larger proportion of the volume of $M^1$, when the class corresponding to $M^1$ is more likely to be confused. Therefore, it is necessary to construct an asymmetric measure for the degree of separation between multiple perceptual manifolds, and we expect this measure to accurately reflect the relative magnitude of the degree of separation.

\begin{algorithm*}[t]
\caption{Pseudocode for The Separation Degree of Perceptual Manifold}
\label{alg4}
%\footnotesize{
\textbf{Input:} Training set $D = \left\{ {\left( {{x_i},{y_i}} \right)} \right\}_{i = 1}^M$ with the total number $C$ of classes. A CNN $\{f(x,\theta_1),g(z,\theta_2)\}$, where $f(\cdot)$ and $g(\cdot)$ denote the feature sub-network and classifier, respectively. \\
\textbf{Output:} The volume of all data manifolds. 
\begin{algorithmic}[1] %[1] enables line numbers
\FOR{$j=1$ to $C$}
\STATE Select the sample set ${D_j} = \left\{ {\left( {{x_i},{y_i}} \right)} \right\}_{i = 1}^{{m_j}}$ for class $j$ from $D$, ${m_j}$ is the number of samples for class $j$.
\STATE Calculate the feature embedding $Z_j=\{z_i \mid z_i = f(x_i,\theta_1)\}_{i=1}^{m_j}$ of $D_j$, ${Z_j} = \left[ {{z_1},{z_2}, \ldots ,{z_{{m_j}}}} \right] \in {\mathbb{R}^{p \times {m_j}}}$.
\ENDFOR
 \STATE There exist $C$ perceptual manifolds $\{M^i \}_{i=1}^{C}$, which consist of point sets $\{Z_i=[z_{i,1},\dots,z_{i,m_i}]\in \mathbb{R}^{p\times m_i} \}_{i=1}^{C}$. Let $Z=[Z_1,\dots,Z_C]\in \mathbb{R}^{p\times {\textstyle \sum_{j=1}^{C}}m_j }$. \\
\FOR{$i=1$ to $C$}
   \STATE Let $Z'=[Z_1,\dots,Z_{i-1},Z_{i+1},\dots,Z_C]\in \mathbb{R}^{p\times (({\textstyle \sum_{j=1}^{C}}m_j)-m_i)}$.
   \STATE Calculate the degree of separation $S(M^i)=\log_{\delta }\det((I+\frac{Z'Z'^T}{{\textstyle \sum_{j=1,j\neq i}^{C}}m_j})^{-1}(I+\frac{ZZ^T}{{\textstyle \sum_{j=1}^{C}}m_j})), \delta = \det(I+\frac{1}{m}Z_iZ_i^T)$ for perceptual manifold $M^i$.
   \ENDFOR
\end{algorithmic}
\end{algorithm*}

Suppose there are $C$ perceptual manifolds $\{M^i \}_{i=1}^{C}$, which consist of point sets $\{Z_i=[z_{i,1},\dots,z_{i,m_i}]\in \mathbb{R}^{p\times m_i} \}_{i=1}^{C}$. Let $Z=[Z_1,\dots,Z_C]\in \mathbb{R}^{p\times {\textstyle \sum_{j=1}^{C}}m_j }$, $Z'=[Z_1,\dots,Z_{i-1},Z_{i+1},\dots,Z_C]\in \mathbb{R}^{p\times (({\textstyle \sum_{j=1}^{C}}m_j)-m_i)}$, we define the degree of separation between the perceptual manifold $M^i$ and the rest of the perceptual manifolds as $$S(M^i)=\frac{V\!ol(Z)-V\!ol(Z')}{V\!ol(Z_i)}.$$

The following analysis is performed for the case when $C=2$ and $V\!ol(Z_2)>V\!ol(Z_1)$. According to our motivation, the measure of the degree of separation between perceptual manifolds should satisfy $S(M^2)>S(M^1)$. 

If $S(M^2)>S(M^1)$ holds, then we can get 
\begin{small}
\begin{equation}
\begin{split}
&V\!ol(Z)V\!ol(Z_1)-V\!ol(Z_1)^2  > V\!ol(Z)V\!ol(Z_2)-V\!ol(Z_2)^2, \\
& \iff V\!ol(Z)(V\!ol(Z_1)-V\!ol(Z_2)) > V\!ol(Z_1)^2-V\!ol(Z_2)^2, \\
& \iff V\!ol(Z)<V\!ol(Z_1)+V\!ol(Z_2).
\nonumber
\end{split}
\end{equation}
\end{small}
We prove that $V\!ol(Z)<V\!ol(Z_1)+V\!ol(Z_2)$ holds when $V\!ol(Z_2)>V\!ol(Z_1)$, and the details are as follows. 

\begin{proof}
Since the function $\log_2\det(\cdot)$ is strictly concave, the real symmetric positive definite matrices $I+\frac{1}{m} Z^TZ$ and $I+\frac{1}{m}diag\{Z_1^TZ_1,Z_2^TZ_2\}$ satisfy \cite{paper57}
\begin{small}
\begin{equation}
\begin{split}
\log_2\det(I+\frac{1}{m}Z^TZ)\le \log_2\det(I+\frac{1}{m}diag\{Z_1^TZ_1,Z_2^TZ_2\}) \\ + tr((I+\frac{1}{m}diag\{Z_1^TZ_1,Z_2^TZ_2\})^T (I+\frac{1}{m}Z^TZ)).
\nonumber
\end{split}
\end{equation}
\end{small}
Also because
%\begin{small}
\begin{equation}
\begin{split}
\log_2\det(I+\frac{1}{m}diag\{Z_1^TZ_1,Z_2^TZ_2\})=\\ \log_2\det(I+\frac{1}{m}Z_1^TZ_1)+\log_2\det(I+\frac{1}{m}Z_2^TZ_2)
\nonumber
\end{split}
\end{equation}
%\end{small}
and
%\begin{small}
\begin{equation}
\begin{split}
& tr((I+\frac{1}{m}diag\{Z_1^TZ_1,Z_2^TZ_2\})^T (I+\frac{1}{m}Z^TZ)) \\ & =tr(diag\{I,I\})=m.
\nonumber
\end{split}
\end{equation}
%\end{small}
We can get 
%\begin{small}
\begin{equation}
\begin{split}
\log_2\det(I+\frac{1}{m}Z^TZ)\le \log_2\det(I+\frac{1}{m}Z_1^TZ_1)\\ +\log_2\det(I+\frac{1}{m}Z_2^TZ_2),
\nonumber
\end{split}
\end{equation}
%\end{small}
i.e., $Vol(Z)<Vol(Z_1)+Vol(Z_2)$ holds.
\end{proof}

The above analysis shows that the proposed measure meets our requirements and motivation. The formula for calculating the degree of separation between perceptual manifolds can be further reduced to 
%\begin{small}
\begin{equation}
\begin{split}
S(M^i)&=\frac{\log_2\det(I+\frac{1}{ {\textstyle \sum_{j=1}^{C}m_j}}ZZ^T)}{\log_2\det(I+\frac{1}{m_i}Z_iZ_i^T)} \\ &-  \frac{\log_2\det(I+\frac{1}{ {\textstyle \sum_{j=1}^{C}m_j}}Z'Z'^T)}{\log_2\det(I+\frac{1}{m_i}Z_iZ_i^T)} \\
&=\frac{\log_2\frac{\det(I+\frac{1}{ {\textstyle \sum_{j=1}^{C}m_j}}ZZ^T)}{\det(I+\frac{1}{ {\textstyle \sum_{j=1,j\neq i}^{C}m_j}}Z'Z'^T)}}{\log_2\det(I+\frac{1}{m_i}Z_iZ_i^T)} \\
&=\log_{\delta }\det((I+\frac{Z'Z'^T}{{\textstyle \sum_{j=1,j\neq i}^{C}}m_j})^{-1}(I+\frac{ZZ^T}{{\textstyle \sum_{j=1}^{C}}m_j})), \\
\delta &= \det(I+\frac{1}{m}Z_iZ_i^T).
\nonumber
\end{split}
\end{equation}
%\end{small}

\begin{figure}[t]
%\vskip -0.05in
\centering
\centerline{\includegraphics[width=1\columnwidth]{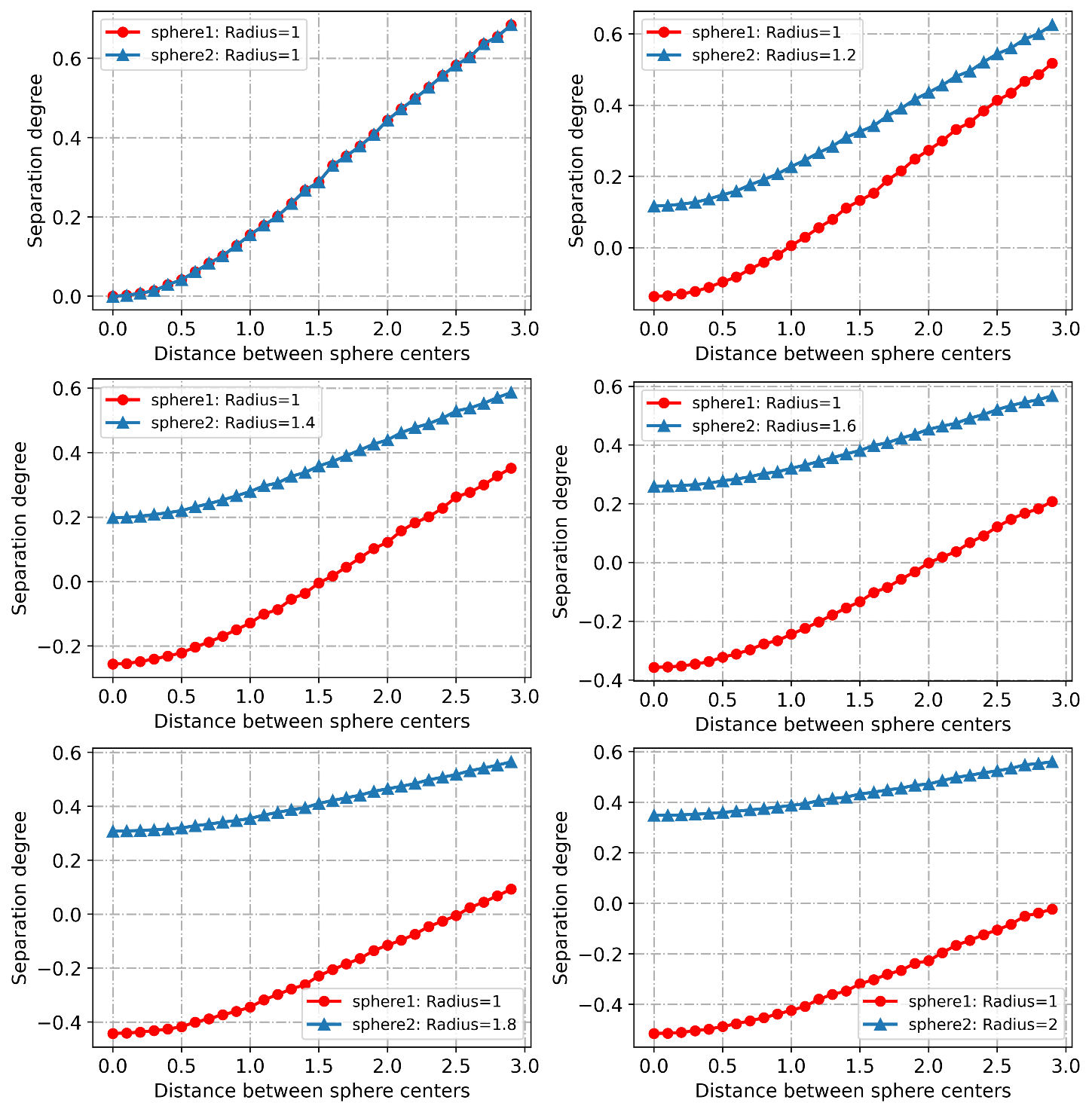}}
\vskip -0.12in
\caption{The variation curve between the separation degree of two spherical point clouds and the distance between spherical centers.}
\label{fig2}
\vskip -0.1in
\end{figure}

Next, we validate the proposed measure of the separation degree between perceptual manifolds in a 3D spherical point cloud scene. Specifically, we conducted the experiments in three cases: 

\begin{itemize}[]
\item[(1)] Construct two 3D spherical point clouds of radius $1$, and then increase the distance between their spherical centers. Since the volumes of the two spherical point clouds are equal, their separation degrees should be symmetric. The variation curves of the separation degrees are plotted in Fig.\ref{fig2}, and it can be seen that the experimental results satisfy our theoretical predictions.

\item[(2)] Change the distance between the centers of two spherical point clouds. Observe their separation degrees, the separation degrees of these two spherical point clouds should be asymmetric. Fig.\ref{fig2} shows that their separation degrees increase as the distance between their centers increases. Also, the manifold with a larger radius has a greater separation degree, and this experimental result conforms to our analysis and motivation.

\item[(3)] As shown in Fig.\ref{fig2}, as the volume difference between the two spherical point cloud manifolds becomes larger, the difference in separation between the two increases, a result that is entirely consistent with our motivation.
\end{itemize}

The separation degree between perceptual manifolds may affect the model's bias towards classes. In addition, it can also be used as the regularization term of the loss function or applied in contrast learning to keep the different perceptual manifolds away from each other.

\subsection{The Curvature of Perceptual Manifold}
\label{sec3.4}

Given a point cloud perceptual manifold $M$, which consists of a $p$-dimensional point set $\{z_1,\dots,z_n\}$, our goal is to calculate the Gauss curvature at each point. First, the normal vector at each point on $M$ is estimated by the neighbor points. Denote by $z_i^j$ the $j$-th neighbor point of $z_i$ and $u_i$ the normal vector at $z_i$. We solve for the normal vector by minimizing the inner product of $z_i^j-c_i,j=1,\dots,k$ and $u_i$ \cite{paper53}, i.e., $$\min {\textstyle \sum_{j=1}^{k}}((z_i^j-c_i)^Tu_i)^2,$$ where $c_i=\frac{1}{k}{\textstyle \sum_{j=1}^{k}}z_i^j$ and $k$ is the number of neighbor points. Let $y_j=z_i^j-c_i$, then the optimization objective is converted to
\begin{equation}
\begin{split}
\min{\textstyle \sum_{j=1}^{k}}(y_j^Tu_i)^2& =\min {\textstyle \sum_{j=1}^{k}}u_i^Ty_jy_j^Tu_i \\
& =\min(u_i^T( {\textstyle \sum_{j=1}^{k}}y_jy_j^T)u_i).
\nonumber
\end{split}
\end{equation}
${\textstyle \sum_{j=1}^{k}}y_jy_j^T$ is the covariance matrix of $k$ neighbors of $z_i$. Therefore, let $Y=[y_1,\dots,y_k]\in \mathbb{R}^{p\times k}$ and ${\textstyle \sum_{j=1}^{k}}y_jy_j^T=YY^T$. The optimization objective is further equated to
\begin{equation}
\begin{split}
\begin{cases} f(u_i)=u_i^TYY^Tu_i,YY^T\in \mathbb{R}^{p\times p},
 \\ min(f(u_i)),
 \\ s.t. u_i^Tu_i=1.
\end{cases} 
\nonumber
\end{split}
\end{equation}
Construct the Lagrangian function $L(u_i,\lambda)=f(u_i)-\lambda (u_i^Tu_i-1)$ for the above optimization objective, where $\lambda$ is a parameter. The first-order partial derivatives of $L(u_i,\lambda)$ with respect to $u_i$ and $\lambda$ are
\begin{equation}
\begin{split}
\frac{\partial L(u_i,\lambda)}{\partial u_i}& =\frac{\partial}{\partial u_i}f(u_i)-\lambda\frac{\partial}{\partial u_i}(u_i^Tu_i-1)  \\
& =2(YY^Tu_i-\lambda u_i),  \\
\frac{\partial L(u_i,\lambda)}{\partial \lambda} & =u_i^Tu_i-1.
\nonumber
\end{split}
\end{equation}
Let $\frac{\partial L(u_i,\lambda)}{\partial u_i}$ and $\frac{\partial L(u_i,\lambda)}{\partial \lambda}$ be $0$, we can get $YY^Tu_i=\lambda u_i,u_i^Tu_i=1$. It is obvious that solving for $u_i$ is equivalent to calculating the eigenvectors of the covariance matrix $YY^T$, but the eigenvectors are not unique. From $\left \langle YY^Tu_i,u_i \right \rangle =\left \langle \lambda u_i,u_i \right \rangle$ we can get $\lambda=\left \langle YY^Tu_i,u_i \right \rangle=u_i^TYY^Tu_i $, so the optimization problem is equated to $\mathop{\arg\min}_{u_i}(\lambda)$. Performing the eigenvalue decomposition on the matrix $YY^T$ yields $p$ eigenvalues $\lambda_1,\dots,\lambda_p$ and the corresponding $p$-dimensional eigenvectors $[\xi_1,\dots,\xi_p]\in \mathbb{R}^{p\times p}$, where $\lambda_1\ge \dots \ge \lambda_p\ge 0$, $\left \| \xi_i \right \| _2=1,i=1,\dots,p$, $\left \langle \xi_a,\xi_b \right \rangle =0(a\neq b)$. The eigenvector $\xi_{m+1}$ corresponding to the smallest non-zero eigenvalue of the matrix $YY^T$ is taken as the normal vector $u_i$ of $M$ at $z_i$.

\begin{algorithm*}[!t]
\caption{Pseudocode for the Mean Gaussian Curvature of The Perceptual Manifold}
\label{alg5}
%\footnotesize{
\textbf{Input:} Given a point cloud perceptual manifold $M$, which consists of a $p$-dimensional point set $\{z_1,\dots,z_n\}$. Denote by $z_i^j$ the $j$-th neighbor point of $z_i$ and $u_i$ the normal vector at $z_i$. \\
\textbf{Output:} The mean Gaussian curvature of the perceptual manifold $M$. 
\begin{algorithmic}[1] %[1] enables line numbers
\FOR{$i=1$ to $n$}
\STATE Select $k$ neighbor points $z_i^j, j=1,\dots,k$ of $z_i$ and let $Y=[z_i,z_i^1,\dots,z_i^k]\in \mathbb{R}^{p\times k}$.
\STATE ${Y} = {Y} - N\!um\!P\!y.mean\left( {{Y}, {\rm{1}}} \right)$.
\STATE Calculate the local covariance matrix $\frac{1}{k}YY^T$.
\STATE Diagonalize $\frac{1}{k}YY^T$ as $U^TDU$ with $D=diag(\lambda_1,\dots,\lambda_p), \lambda_1\ge \dots \ge \lambda_{m+1} >  \lambda_{m+2}=\dots=0, U=[\xi_1,\dots,\xi_p]\in \mathbb{R}^{p\times p}, \left \| \xi_i \right \| _2=1,i=1,\dots,p, \left \langle \xi_a,\xi_b \right \rangle =0(a\neq b)$.
\STATE Let $u_i=\lambda_{m+1}$.
\STATE The $k$ neighbors of $z_i$ are projected into the affine space $z_i+\left \langle \xi_1,\dots,\xi_m \right \rangle$ and denoted as $o_j=[(z_i^j-z_i)\cdot \xi_1,\dots,(z_i^j-z_i)\cdot \xi_m]^T\in \mathbb{R}^m,j=1,\dots,k$.
\STATE Denote by $o_j[m]$ the $m$-th component $(z_i^j-z_i)\cdot \xi_m$ of $o_j$. We use $z_i$ and $k$ neighbor points to fit a quadratic hypersurface $f(\theta)$ with parameter $\theta \in \mathbb{R}^{m\times m}$. The hypersurface equation is denoted as $f(o_j,\theta)=\frac{1}{2} {\textstyle \sum_{a,b}}\theta_{a,b}o_j\left [ a \right ] o_j\left [ b \right ] ,j\in \left \{ 1,\dots,k \right \}$.
\STATE Expand the parameter $\theta$ of the hypersurface into the column vector $\theta=\left [ \theta_{1,1},\dots,\theta_{1,m},\theta_{2,1},\dots,\theta_{m,m} \right]^T\in \mathbb{R}^{m^2}.$.
\STATE Organize the $k$ neighbor points $\left \{ o_j \right \}_{j=1}^k $ of $z_i$ according to the following form:
Organize the $k$ neighbor points $\left \{ o_j \right \}_{j=1}^k $ of $z_i$ according to the following form:
$$ O(z_i)=\begin{bmatrix} o_1\left [ 1 \right ] o_1\left [ 1 \right ] 
  &  o_1\left [ 1 \right ] o_1\left [ 2 \right ]  & \cdots  &  o_1\left [ m \right ] o_1\left [ m \right ]  \\
  o_2\left [ 1 \right ] o_2\left [ 1 \right ]  & o_2\left [ 1 \right ] o_2\left [ 2 \right ] & \cdots &o_2\left [ m \right ] o_2\left [ m \right ] \\
 \vdots  & \vdots  & \ddots  & \vdots  \\
 o_k\left [ 1 \right ] o_k\left [ 1 \right] & o_k\left [ 1 \right ] o_k\left [ 2 \right] & \cdots  &o_k\left [ m \right ] o_k\left [ m \right]
\end{bmatrix}\in \mathbb{R}^{k\times m^2}.$$
\STATE The target value is $T=\left [ (z_i^1-z_i)\cdot u_i,(z_i^2-z_i)\cdot u_i,\dots,(z_i^k-z_i)\cdot u_i \right ]^T \in \mathbb{R}^{k}$.
\STATE Solve for $\frac{\partial}{\partial \theta} (\frac{1}{2}tr\left [ \left ( O(z_i)\theta-T \right)^T(O(z_i)\theta-T) \right ])=0$ to get $\theta = (O(z_i)^TO(z_i))^{-1}O(z_i)^TT$.
\STATE The Gauss curvature of the perceptual manifold $M$ at $z_i$ can be calculated as $G(z_i)=det(\theta)=det((O(z_i)^TO(z_i))^{-1}O(z_i)^TT)$.
\ENDFOR
\STATE The average Gaussian curvature $\frac{1}{n} {\textstyle \sum_{i=1}^{n}}G(z_i)$ of the perceptual manifold $M$ is the average of the Gauss curvatures at all points on $M$.
\end{algorithmic}
\end{algorithm*}

\begin{figure*}[t]
\centering
%\vskip -0.05in
\centerline{\includegraphics[width=1.8\columnwidth]{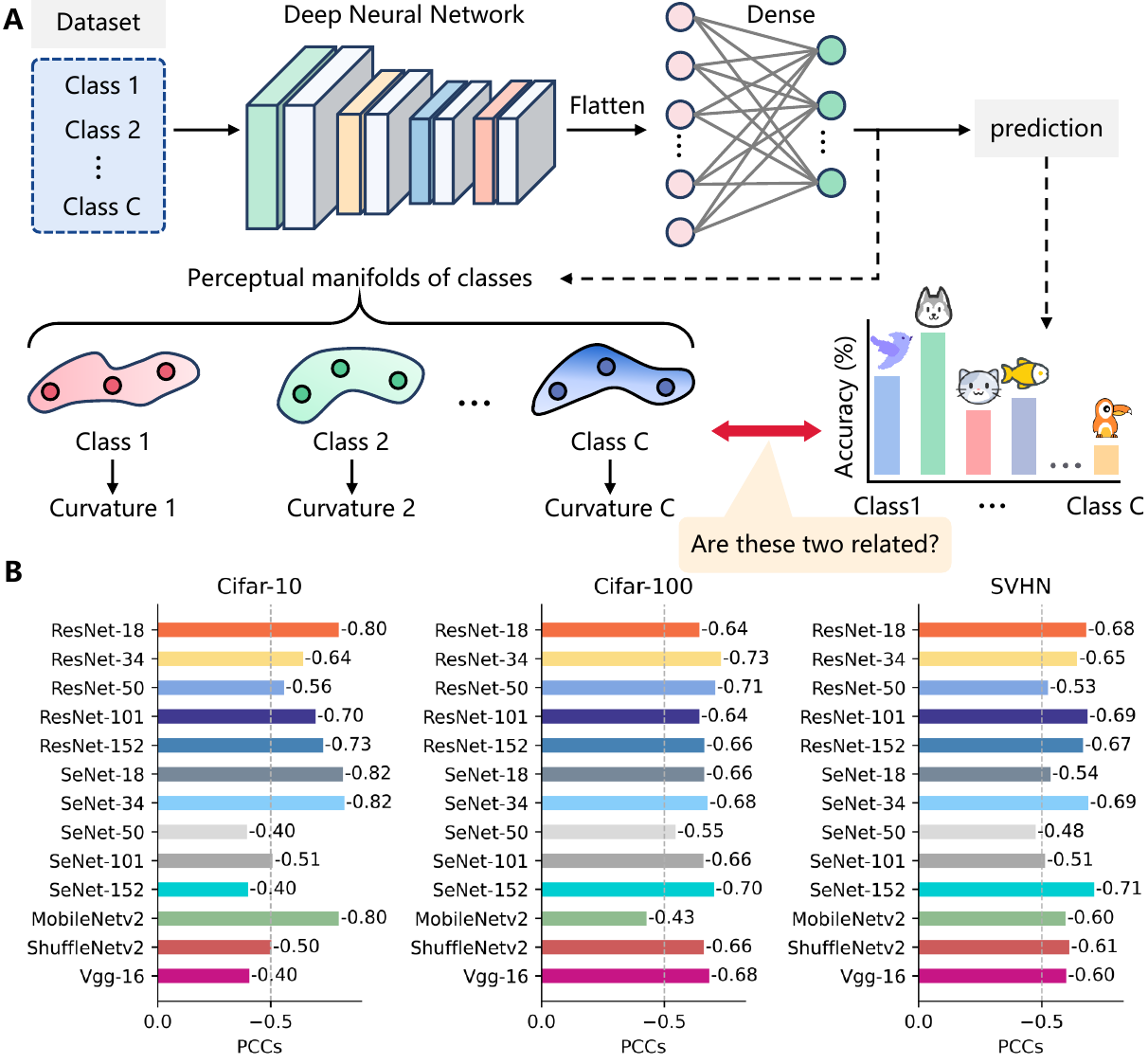}}
\vskip -0.1in
\caption{\textbf{A} The schematic diagram for calculating the curvature of class perceptual manifolds versus class accuracy. \textbf{B} Trained $13$ different models on three sample-balanced datasets, CIFAR-10, CIFAR-100, and SVHN, and calculated the correlation between the curvature of class perceptual manifolds generated by each model and class accuracy. The experimental settings for Figs 4, 6, 8, 9, and 11 are as follows: On the CIFAR-10, CIFAR-100, and SVHN datasets, we used SGD with a momentum of $0.9$, a batch size of $64$, and an initial learning rate of $0.1$. The difference is that on CIFAR-10 and SVHN, the model was trained for $60$ epochs, and the cosine annealing strategy was used for learning rate decay. On CIFAR-100, the model was trained for $200$ epochs, and the learning rate was adjusted to $0.02$, $0.004$, and $0.0008$ at epochs $60$, $120$, and $160$, respectively. All models were trained using Cross-Entropy (CE) loss.}
\label{fig101}
\vskip -0.1in
\end{figure*}

Consider an $m$-dimensional affine space with center $z_i$, which is spanned by $\xi_1,\dots,\xi_m$. This affine space approximates the tangent space at $z_i$ on $M$. We estimate the curvature of $M$ at $z_i$ by fitting a quadratic hypersurface in the tangent space utilizing the neighbor points of $z_i$. The $k$ neighbors of $z_i$ are projected into the affine space $z_i+\left \langle \xi_1,\dots,\xi_m \right \rangle$ and denoted as 
\begin{small}
\begin{equation}
\begin{split}
o_j=[(z_i^j-z_i)\cdot \xi_1,\dots,(z_i^j-z_i)\cdot \xi_m]^T\in \mathbb{R}^m,j=1,\dots,k.
\nonumber
\end{split}
\end{equation}
\end{small}
Denote by $o_j[m]$ the $m$-th component $(z_i^j-z_i)\cdot \xi_m$ of $o_j$. We use $z_i$ and $k$ neighbor points to fit a quadratic hypersurface $f(\theta)$ with parameter $\theta \in \mathbb{R}^{m\times m}$. The hypersurface equation is denoted as
%\begin{small}
\begin{equation}
\begin{split}
f(o_j,\theta)=\frac{1}{2} {\textstyle \sum_{a,b}}\theta_{a,b}o_j\left [ a \right ] o_j\left [ b \right ] ,j\in \left \{ 1,\dots,k \right \}, 
\nonumber
\end{split}
\end{equation}
%\end{small}
further, minimize the squared error
%\begin{small}
\begin{equation}
\begin{split}
E(\theta)= {\textstyle \sum_{j=1}^{k}}( \frac{1}{2} {\textstyle \sum_{a,b}}\theta_{a,b}o_j\left [ a \right ] o_j\left [ b \right ] -(z_i^j-z_i)\cdot u_i)^2.
\nonumber
\end{split}
\end{equation}
%\end{small}
Let $\frac{\partial E(\theta)}{\partial \theta_{a,b}}=0,a,b\in \left \{ 1,\dots,m \right \}$ yield a nonlinear system of equations, but it needs to be solved iteratively. Here, we propose an ingenious method to fit the hypersurface and \textbf{give the analytic solution of the parameter $\theta$} directly. Expand the parameter $\theta$ of the hypersurface into the column vector
%\begin{small}
\begin{equation}
\begin{split}
\theta=\left [ \theta_{1,1},\dots,\theta_{1,m},\theta_{2,1},\dots,\theta_{m,m} \right]^T\in \mathbb{R}^{m^2}.
\nonumber
\end{split}
\end{equation}
%\end{small}
Organize the $k$ neighbor points $\left \{ o_j \right \}_{j=1}^k $ of $z_i$ according to the following form:
\begin{scriptsize}
\begin{equation}
\begin{split}
O(z_i)=\begin{bmatrix} o_1\left [ 1 \right ] o_1\left [ 1 \right ] 
  &  o_1\left [ 1 \right ] o_1\left [ 2 \right ]  & \cdots  &  o_1\left [ m \right ] o_1\left [ m \right ]  \\
  o_2\left [ 1 \right ] o_2\left [ 1 \right ]  & o_2\left [ 1 \right ] o_2\left [ 2 \right ] & \cdots &o_2\left [ m \right ] o_2\left [ m \right ] \\
 \vdots  & \vdots  & \ddots  & \vdots  \\
 o_k\left [ 1 \right ] o_k\left [ 1 \right] & o_k\left [ 1 \right ] o_k\left [ 2 \right] & \cdots  &o_k\left [ m \right ] o_k\left [ m \right]
\end{bmatrix}\in \mathbb{R}^{k\times m^2}.
\nonumber
\end{split}
\end{equation}
\end{scriptsize}
The target value is
\begin{small}
\begin{equation}
\begin{split}
T=\left [ (z_i^1-z_i)\cdot u_i,(z_i^2-z_i)\cdot u_i,\dots,(z_i^k-z_i)\cdot u_i \right ]^T \in \mathbb{R}^{k}.
\nonumber
\end{split}
\end{equation}
\end{small}
We minimize the squared error
%\begin{small}
\begin{equation}
\begin{split}
E(\theta)=\frac{1}{2}tr\left [ \left ( O(z_i)\theta-T \right)^T(O(z_i)\theta-T) \right ],
\nonumber
\end{split}
\end{equation}
%\end{small}
and find the partial derivative of $E(\theta)$ for $\theta$:
\begin{small}
\begin{equation}
\begin{split}
\frac{\partial E(\theta)}{\partial \theta}& =\frac{1}{2} \left ( \frac{\partial tr(\theta^TO(z_i)^TO(z_i)\theta)}{\partial \theta}-\frac{\partial tr(\theta^TO(z_i)^TT)}{\partial \theta}   \right ) \\
& =O(z_i)^TO(z_i)\theta-O(z_i)^TT.
\nonumber
\end{split}
\end{equation}
\end{small}
Let $\frac{\partial E(\theta)}{\partial \theta}=0$, we can get
%\begin{small}
\begin{equation}
\begin{split}
\theta = (O(z_i)^TO(z_i))^{-1}O(z_i)^TT.
\nonumber
\end{split}
\end{equation}
%\end{small}
Thus, the Gauss curvature of the perceptual manifold $M$ at $z_i$ can be calculated as
%\begin{small}
\begin{equation}
\begin{split}
G(z_i)=det(\theta)=det((O(z_i)^TO(z_i))^{-1}O(z_i)^TT).
\nonumber
\end{split}
\end{equation}
%\end{small}

\begin{figure}[!t]
\centering
%\vskip -0.05in
\centerline{\includegraphics[width=1\columnwidth]{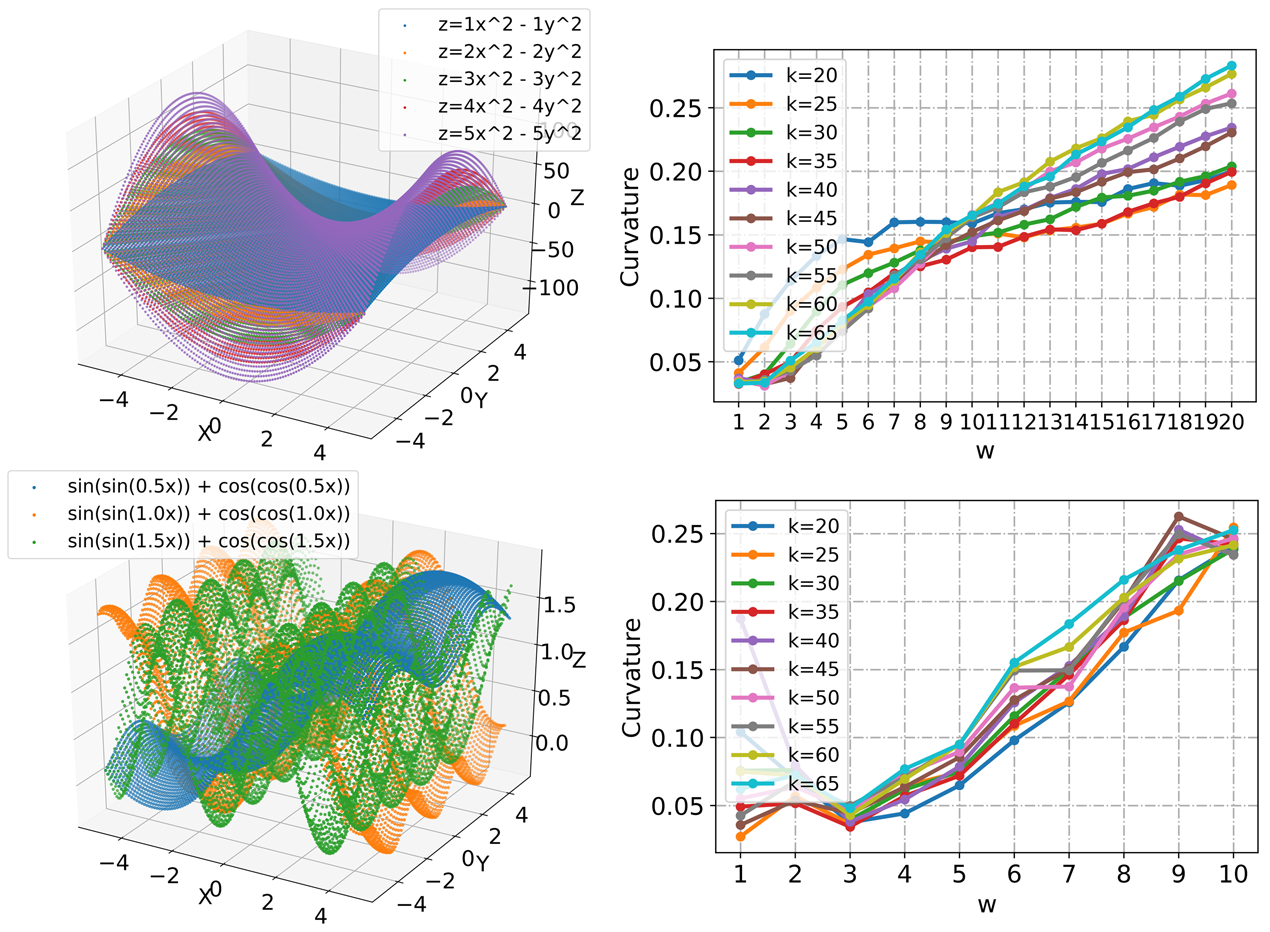}}
\vskip -0.05in
\caption{The surface equations in the first and second rows are $Z=w(X^2-Y^2)$ and $Z=\sin(\sin(0.5wX))+\cos(\cos(0.5wX))$, respectively. We increase the curvature of the surface by increasing $w$ and calculate the complexity of the two-dimensional point cloud surface. Also, we investigate the effect of the number of neighbors $k$ on the complexity of the manifold.}
\label{fig3}
%\vskip -0.1in
\end{figure}

Up to this point, we provide an approximate solution of the Gauss curvature at any point on the point cloud perceptual manifold $M$. \cite{paper56} shows that on a high-dimensional dataset, almost all samples lie on convex locations, and thus the complexity of the perceptual manifold is defined as the average $\frac{1}{n} {\textstyle \sum_{i=1}^{n}}G(z_i)$ of the Gauss curvatures at all points on $M$. Our approach does not require iterative optimization and can be quickly deployed in a deep neural network to calculate the Gauss curvature of the perceptual manifold. Taking the two-dimensional surface in Fig.\ref{fig3} as an example, the surface complexity increases as the surface curvature is artificially increased. This indicates that our proposed complexity measure of perceptual manifold can accurately reflect the changing trend of the curvature degree of the manifold. In addition, Fig.\ref{fig3} shows that the selection of the number of neighboring points hardly affects the monotonicity of the complexity of the perceptual manifold. In our work, we select the number of neighboring points to be $40$.

\begin{figure}[t]
%\vskip -0.05in
\centering
\centerline{\includegraphics[width=0.97\columnwidth]{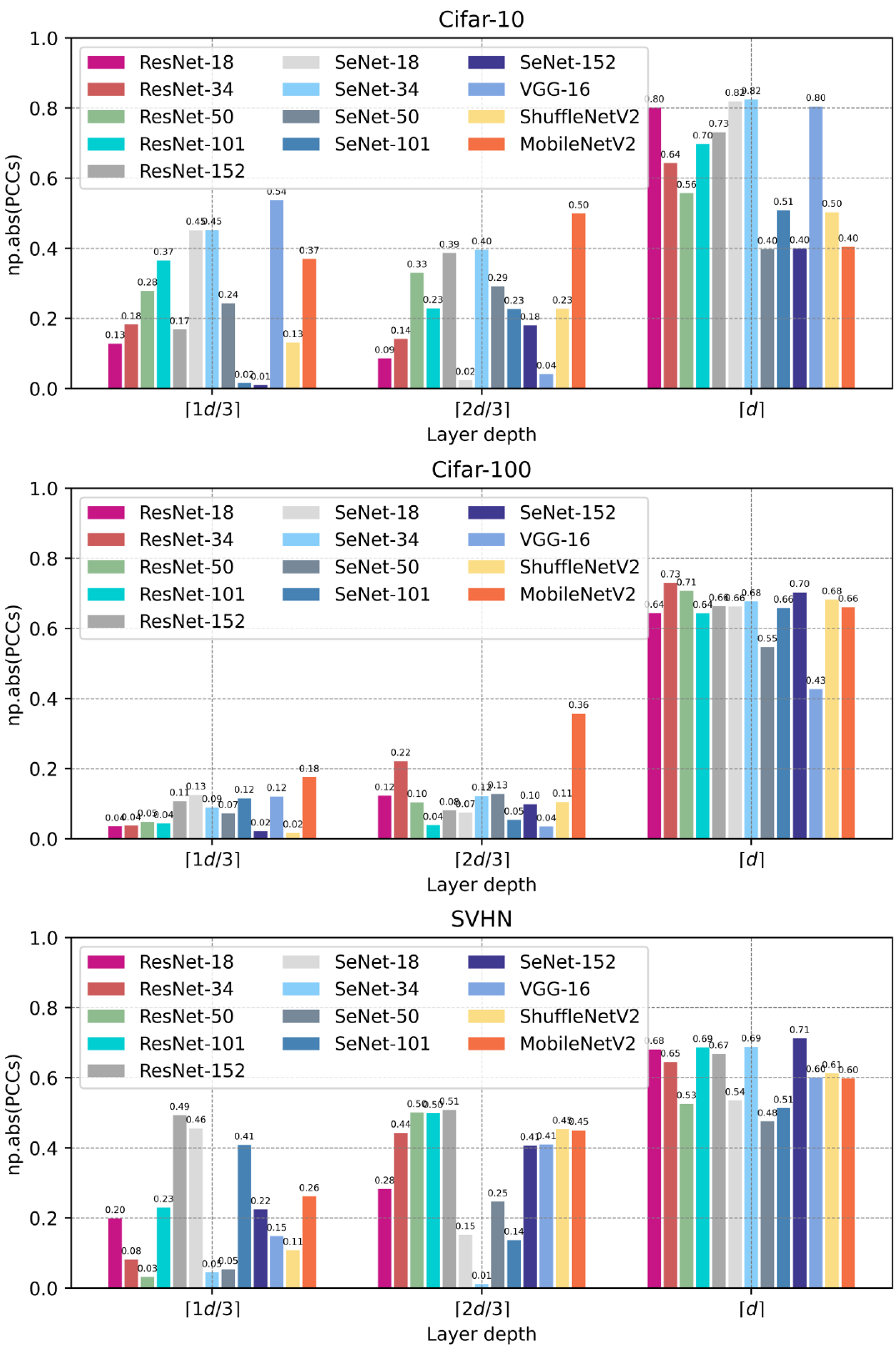}}
\vskip -0.1in
\caption{Absolute values of the Pearson correlation coefficients between the curvature of the class perceptual manifolds generated by the different layers of the deep neural network and the class accuracy. Please note that all correlations in the figure are negative, and absolute values are used to avoid inversion of the bar chart.}
\label{fig102}
\vskip -0.05in
\end{figure}

\begin{figure}[!t]
%\vskip -0.1in
\centering
\centerline{\includegraphics[width=0.97\columnwidth]{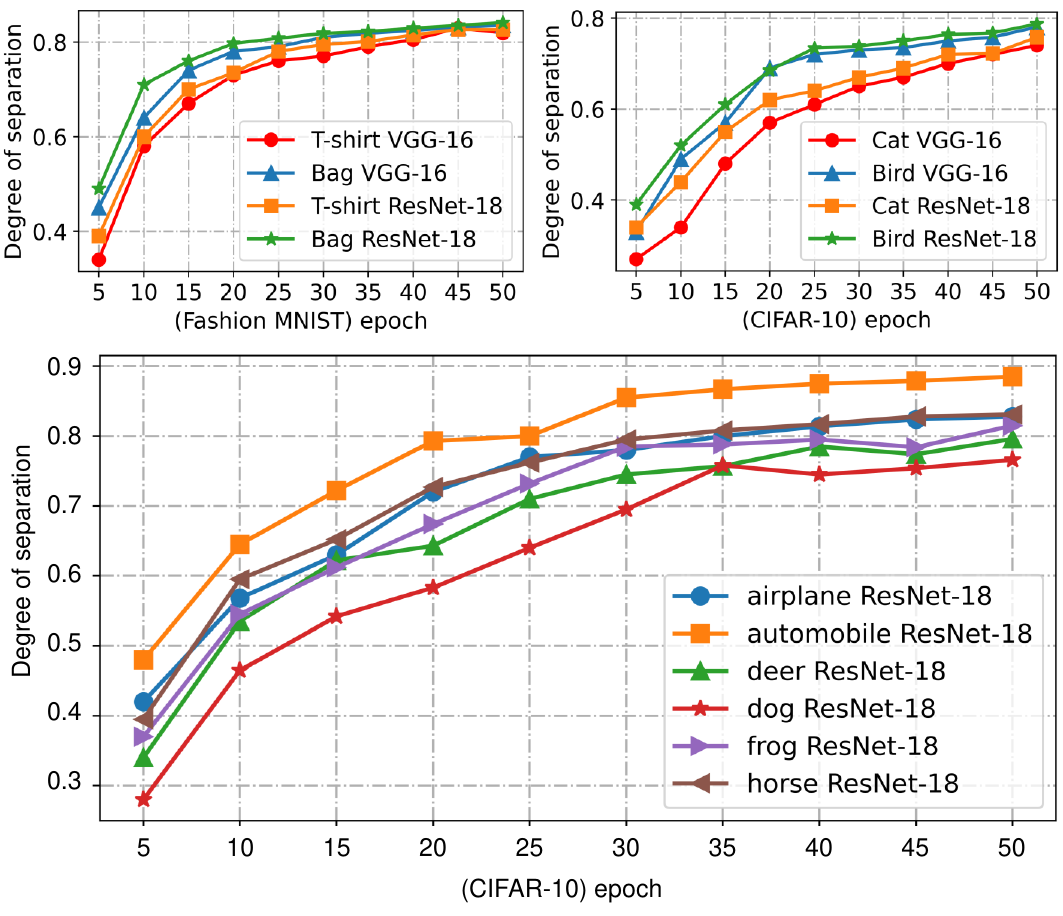}}
\vskip -0.10in
\caption{The variation curves between the separation degree of perceptual manifolds and training epochs on both datasets.}
\label{fig4}
\vskip -0.05in
\end{figure}

\section{The curvature of class perceptual manifolds can predict model bias}
\label{newsec4}

The data manifold is gradually reduced and compressed along the layers of deep neural networks for ease of classification. Intuitively, we speculate that if the curvature of the perceptual manifold of a certain class generated at the last hidden layers of a deep neural network is larger, the difficulty of classifying that class will also increase. When a model exhibits inconsistency across classes, it is typically considered biased. In this section, we comprehensively explore the relationship between class perception manifold curvature and model bias.

As shown in Fig.\ref{fig101}, we first extract image embeddings corresponding to each class generated by the last hidden layer of a well-trained DNN, forming class perceptual manifolds. Subsequently, we estimate the curvature of each perceptual manifold and compute the Pearson correlation coefficient between curvature and class accuracy. Experimental results are presented in Fig.\ref{fig101}, revealing a significant negative correlation between the curvature of class perceptual manifolds and class accuracy across three datasets with balanced sample sizes. Particularly on CIFAR-100, $13$ models consistently demonstrate a pronounced negative correlation, suggesting the universality of our findings. Given the low probability of such results occurring by chance on a dataset with $100$ classes, this discovery not only offers a new tool for investigating model fairness but also underscores the vast potential in analyzing the behavior of deep neural networks from a geometric perspective.

Furthermore, we were curious whether the curvature of class perceptual manifolds generated by other layers in deep neural networks could predict model bias. Assuming the number of layers in the model is $d$, in addition to extracting image embeddings at the last hidden layer, we also extract image embeddings at the $\lceil 1d/3 \rceil$ and $\lceil 2d/3 \rceil$ layers. Similar to Fig.\ref{fig101}, we compute the Pearson correlation coefficient between the curvature of class perceptual manifolds and class accuracy and present the results in Fig.\ref{fig102}. We found that only the curvature of perceptual manifolds generated at the last hidden layer can reliably predict model bias. This phenomenon is particularly pronounced on datasets with a large number of classes, such as CIFAR-100. In summary, we have discovered a new mechanism for explaining model fairness and it is possible that it could serve as a geometric constraint to make models fairer.

\section{How Learning Shapes the Geometric Characteristics of Perceptual Manifolds}
\label{sec4}

Our experiments have demonstrated that, when a deep neural network is well-trained, the curvature of class perceptual manifolds generated by its last hidden layer can predict its bias toward classes. This finding suggests that existing models may struggle to handle biases introduced by curvature imbalances during the learning process. In contrast, we presume that existing models may effectively reduce the correlation between the separability of perceptual manifolds and model bias, as intuitively separability is a fundamental goal in classification tasks. In the following, we systematically explore how learning affects the geometric properties of perceptual manifolds.

\begin{figure*}[t]
%\vskip -0.1in
\centering
\centerline{\includegraphics[width=2.05\columnwidth]{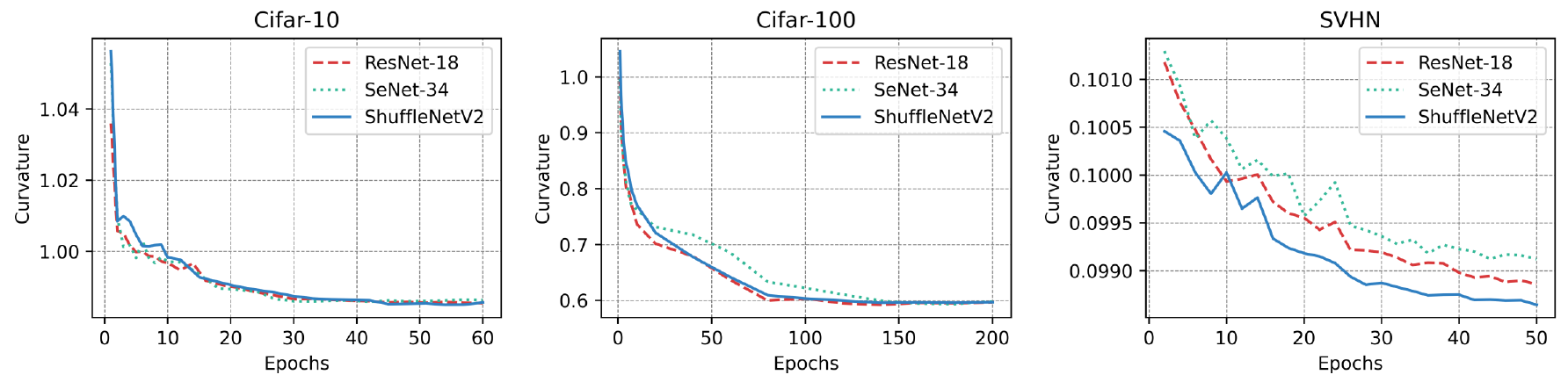}}
\vskip -0.1in
\caption{The variation curves between the curvature of perceptual manifolds and training epochs on both datasets.}
\label{fig5}
%\vskip -0.2in
\end{figure*}

\subsection{Learning Facilitates the Separation}
\label{sec4.1}

Learning typically leads to greater inter-class distance, which equates to greater separation between perceptual manifolds. We trained VGG-16 \cite{paper16} and ResNet-18 \cite{paper21} on F-MNIST \cite{paper55} and CIFAR-10 \cite{paper25} to explore the effect of the learning process on the separation degree between perceptual manifolds and observed the following phenomenon.

As shown in Fig.\ref{fig4}, each perceptual manifold is gradually separated from the other manifolds during training. It is noteworthy that the separation is faster in the early stage of training, and the increment of separation degree gradually decreases in the later stage.

\begin{figure}[!t]
%\vskip -0.1in
\centering
\centerline{\includegraphics[width=1.02\columnwidth]{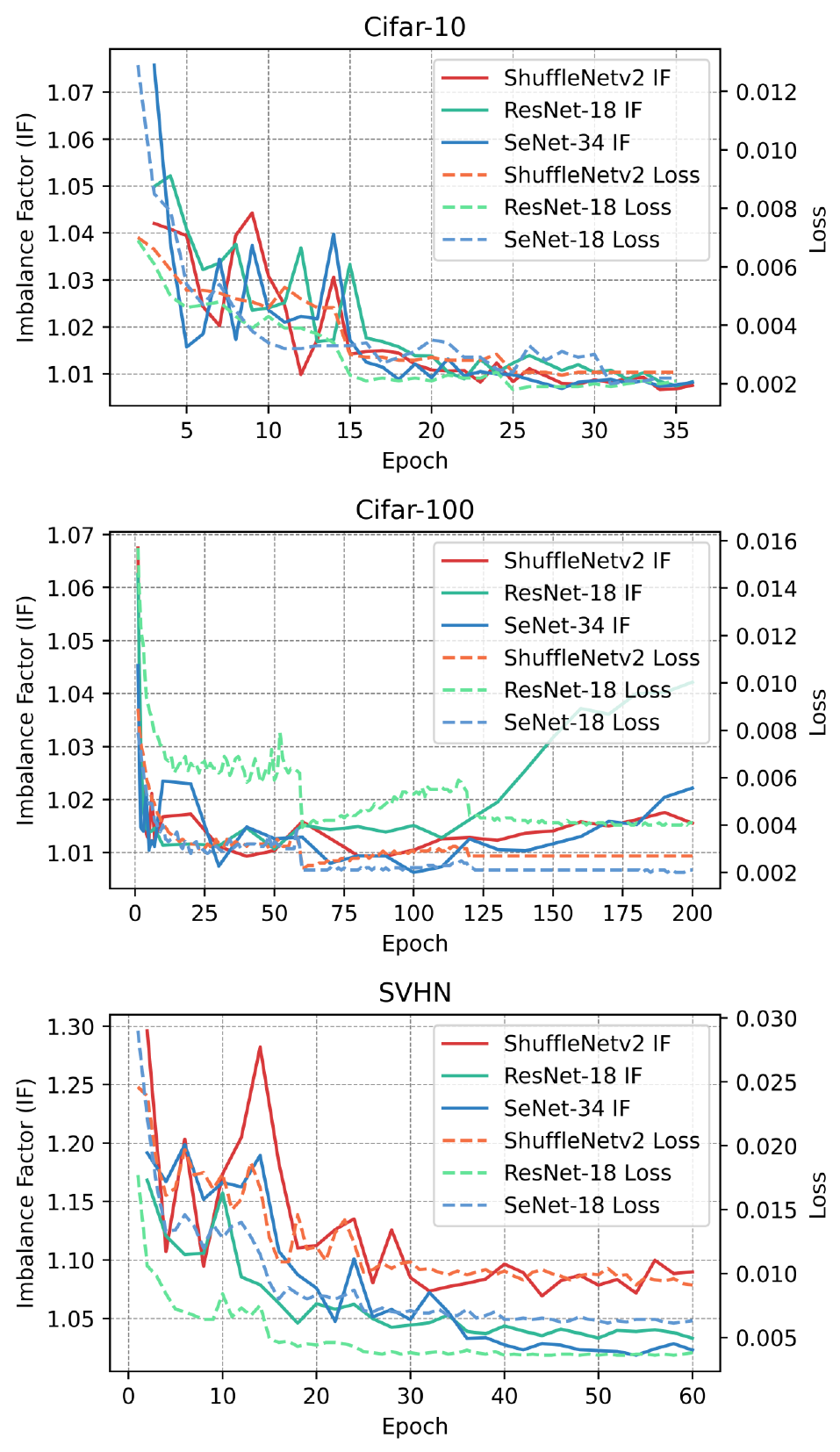}}
\vskip -0.1in
\caption{Curves of loss and degree of curvature imbalance with epoch.}
\label{fig103}
%\vskip -0.2in
\end{figure}

\subsection{Learning Reduces Curvature and Its Imbalance}
\label{sec4.2}

We conducted experiments on CIFAR-10, CIFAR-100, and SVHN by training ResNet-18, SeNet-34, and ShuffleNetV2 models. We extracted embeddings for each class of images at different training stages of the models. For visualization purposes, we averaged the curvature of perceptual manifolds corresponding to each class and plotted them in Fig.\ref{fig5}. It can be observed that the curvature of perceptual manifolds decreases rapidly in the early stages of training, but as training progresses, the rate of decrease gradually slows down. Compared to the initial curvature, the degree of decrease appears less pronounced, for example, the curvature of perceptual manifolds decreased by less than $10\%$ on CIFAR-10. We speculate that this is due to the lack of curvature constraints in the optimization objective, leading to the rapid initial decrease because of the widespread information compression ability in deep neural networks. To confirm this hypothesis, we plotted the curves of loss decrease and curvature imbalance as a function of epochs in Fig.\ref{fig103}. It can be observed that on all three datasets, as the loss gradually converges, the rate of decrease in curvature and imbalance also gradually decreases.

The above experiments indicate that deep neural networks, driven by optimization objectives lacking curvature constraints, are still capable of reducing the curvature of perceptual manifolds through information compression. This is understandable, as without information compression, achieving classification would be challenging. However, we must consider whether optimization objectives without curvature constraints are sufficient to address model bias caused by curvature imbalance. In the next subsection, we visualize the curve of the correlation between the curvature of the class perceptual manifold and the class accuracy with epoch to answer this question.

\subsection{Curvature Imbalance and Model Bias}
\label{sec4.3}

Although existing models separate class perceptual manifolds from each other during the learning process and also flatten the perceptual manifolds, do existing models have enough power to adequately mitigate the model bias caused by these two factors? We trained VGG-16 and ResNet-18 on Fashion MNIST and CIFAR-10 and plotted the correlation between the separation and curvature of class perceptual manifolds and class accuracy as a function of epoch.

\begin{figure}[h]
\centering
%\vskip -0.05in
\centerline{\includegraphics[width=1\columnwidth]{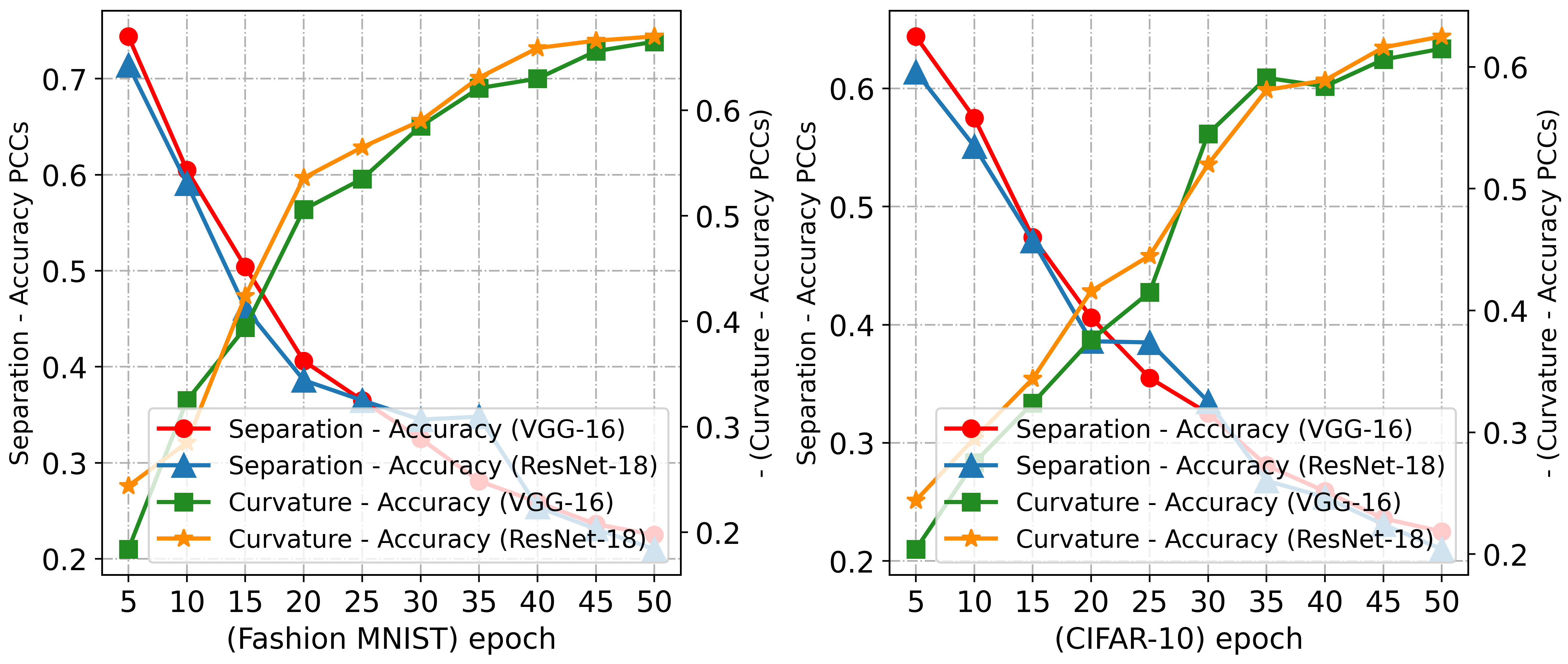}}
\vskip -0.05in
\caption{The Pearson correlation coefficients (PCCs) between the accuracy of all classes and the separation degree and curvature of the corresponding perceptual manifolds, respectively.}
\label{fig6}
%\vskip -0.07in
\end{figure}

\begin{figure}[h]
\vskip -0.05in
\centering
\centerline{\includegraphics[width=1\columnwidth]{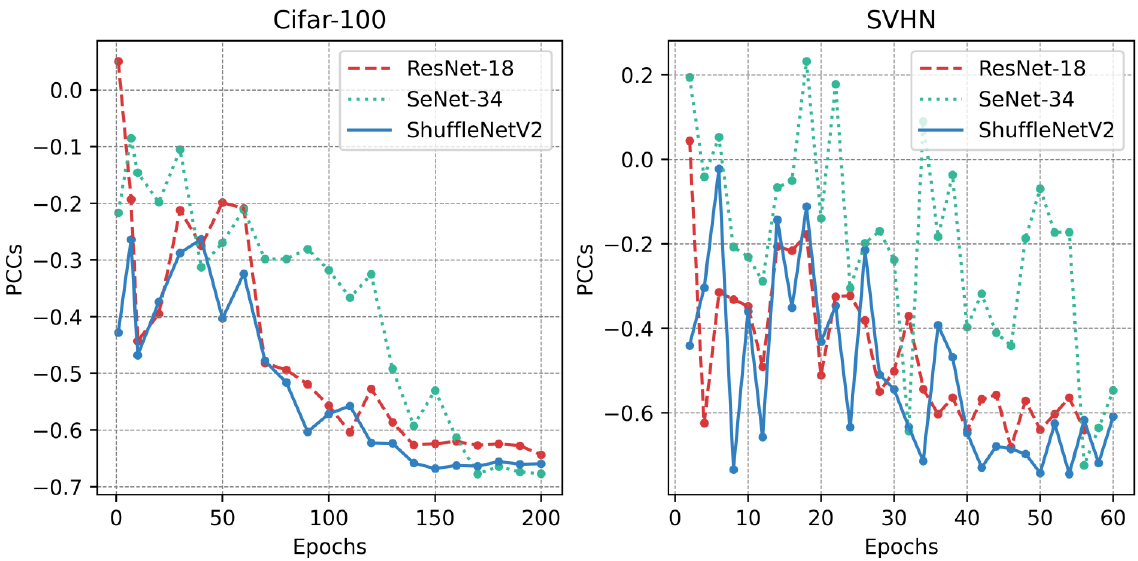}}
\vskip -0.1in
\caption{Curves of the PCCs between the curvature of the class perceptual manifold and the class accuracy with epoch.}
\label{fig104}
%\vskip -0.2in
\end{figure}

The experimental results, as shown in Fig.\ref{fig6}, reveal a decrease in the correlation between the separability of perceptual manifolds and the accuracy of corresponding classes as training progresses, while the negative correlation between curvature and accuracy increases. This suggests that existing methods can only alleviate the impact of the separability between perceptual manifolds on model bias while overlooking the influence of the complexity of perceptual manifolds on model bias. Additionally, we further trained ResNet-18, SeNet-34, and ShuffleNetV2 on CIFAR-100 and SVHN to thoroughly observe the trends of curvature and class accuracy of class-aware manifolds as a function of epochs (see Fig.\ref{fig104}). The experimental results demonstrate that existing models lack constraints on curvature during the training process, resulting in a highly negative correlation between curvature and class accuracy after model convergence.

%\vspace{-3pt}
\section{Curvature-Balanced Feature Learning}
\label{sec5}
%\vspace{-3pt}

The above study shows that it is necessary to focus on the model bias caused by the curvature imbalance among perceptual manifolds. In this section, we propose curvature regularization, which can reduce the model bias and further improve the performance of existing methods.

\subsection{Design Principles of The Proposed Approach}
\label{sec5.1}

The proposed curvature regularization needs to satisfy the following three principles to learn curvature-balanced and flat perceptual manifolds.

\textbf{(1)} The greater the curvature of a perceptual manifold, the stronger the penalty for it. Our experiments show that learning reduces the curvature, so it is reasonable to assume that flatter perceptual manifolds are easier to decode.
\textbf{(2)} When the curvature is balanced, the penalty strength is the same for each perceptual manifold.
\textbf{(3)} The sum of the curvatures of all perceptual manifolds tends to decrease.

\iffalse
\begin{itemize}
    \item[(1)] The greater the curvature of a perceptual manifold, the stronger the penalty for it. Our experiments show that learning reduces the curvature, so it is reasonable to assume that flatter perceptual manifolds are easier to decode.
    \item[(2)] When the curvature is balanced, the penalty strength is the same for each perceptual manifold.
     \item[(3)] The sum of the curvatures of all perceptual manifolds tends to decrease.
\end{itemize}
\fi

\subsection{Curvature Regularization (CR)}
\label{sec5.2}

In order to propose curvature regularization in a reasonable way, we start from softmax cross-entropy loss to inspire our method. Given a $C$ classification task, suppose a sample $x$ is labeled as $Y_k$ and it is predicted as each class with probabilities $P_1,P_2,\dots,P_C$, respectively. The cross-entropy loss generated by sample $x$ is calculated as $L(x)= {\textstyle \sum_{i=1}^{C}}-Y_i\log(P_i)$, where $Y_k=1, Y_i=0, i \neq k$. The goal of $L(x)$ is to make $\log(P_k)$ converge to $0$, i.e., $P_k$ converges to $1$, at which point $P_i (i\neq k)$ converges to $0$. Unlike cross-entropy loss, which can pull apart the difference between $P_k$ and other probabilities, we expect the mean Gaussian curvature of the $C$ perceptual manifolds to converge to equilibrium.

\begin{figure}[h]
%\vskip -0.1in
\centering
\centerline{\includegraphics[width=1\columnwidth]{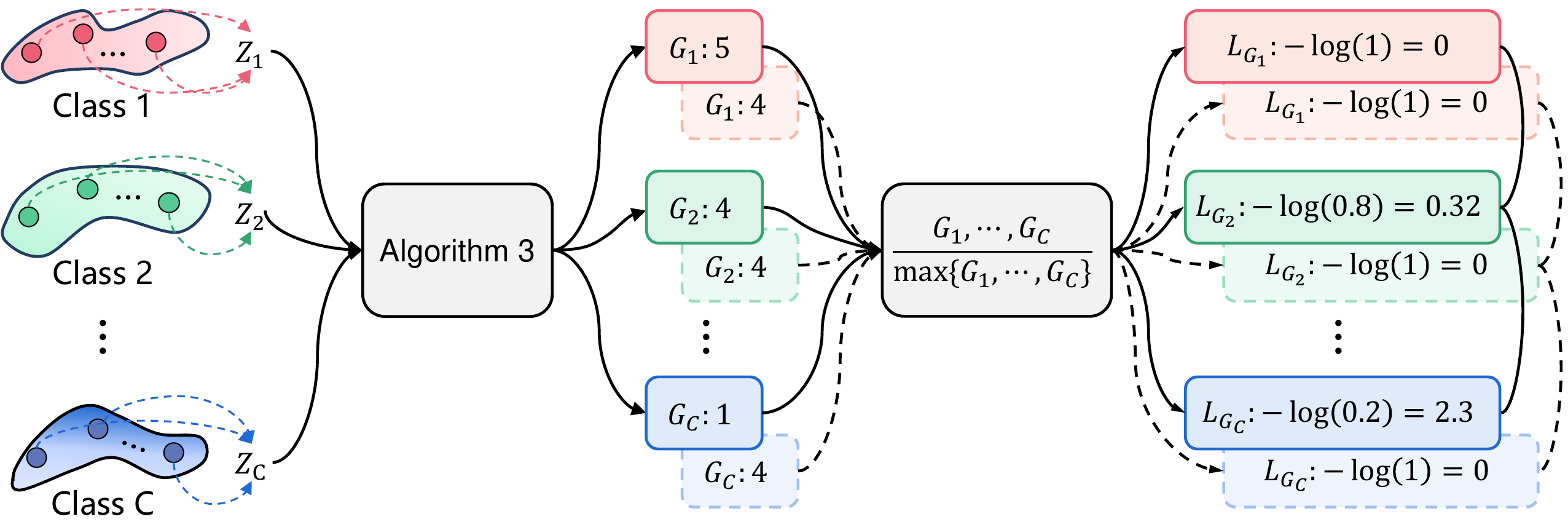}}
%\vskip -0.05in
\caption{All curvatures smaller than $G_1$ gradually increase driven by the loss function, and the smaller the curvature, the greater the resulting loss. $Z_i$ denotes the set of image embeddings of class $i$.}
\label{fig14}
%\vskip -0.12in
\end{figure}

\begin{figure}[h]
%\vskip -0.1in
\centering
\centerline{\includegraphics[width=1\columnwidth]{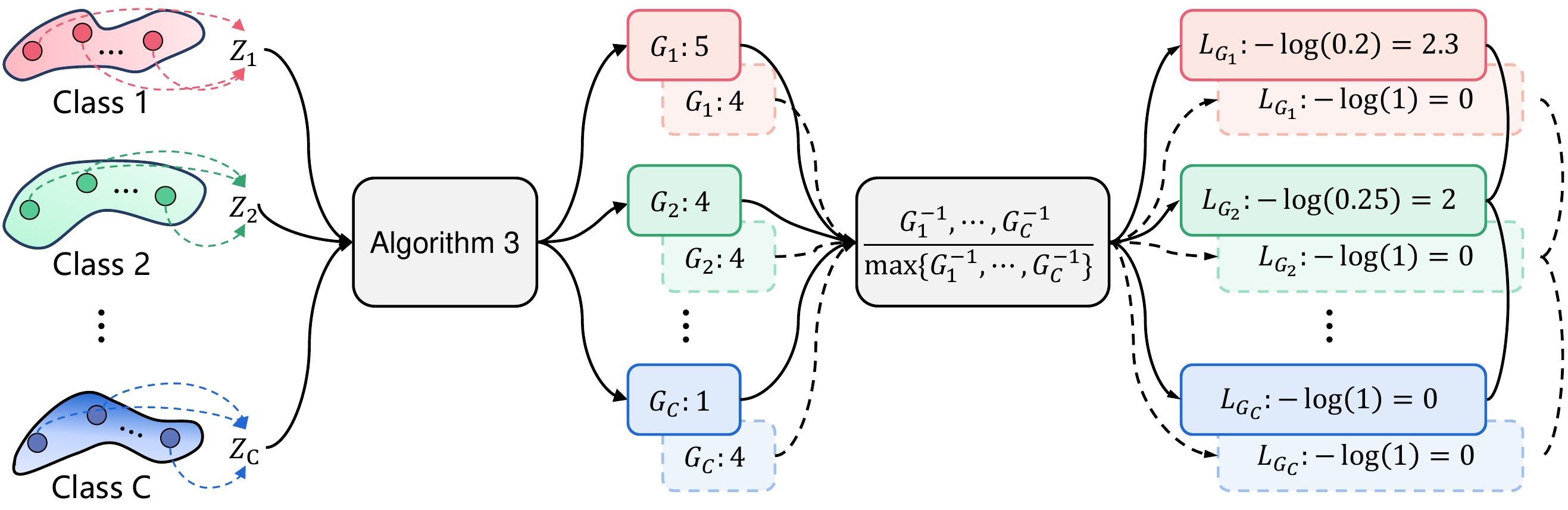}}
%\vskip -0.05in
\caption{The perceptual manifold with the smallest curvature produces no loss, and the larger the curvature, the larger the loss.}
\label{fig15}
%\vskip -0.1in
\end{figure}

Assume that the mean Gaussian curvatures of the $C$ perceptual manifolds are $G_1,G_2,\dots,G_C$ (Algorithm \ref{alg5}), and perform the maximum normalization on them. The $-\log(G_k)$ loss can make $G_k$ converge to $1$. Therefore, perform a negative logarithmic transformation on the curvature of all perceptual manifolds and use it as loss, which can make each curvature converge to $1$ and thus achieve curvature balance. However, the above operation violates the third design principle of curvature regularization, which is that the sum of curvatures of all perceptual manifolds tends to decrease. As shown in Fig.\ref{fig14}, all curvatures smaller than $G_1$ gradually increase driven by the loss function, and the smaller the curvature, the greater the resulting loss. To solve this problem, we update each curvature to the inverse of itself before performing the maximum normalization of the curvature. Eventually, the curvature penalty term of the perceptual manifold $M^i$ is denoted as $-\log(\frac{G_i^{-1}}{\max\{G_1^{-1},\dots,G_C^{-1} \}} )$. Considering the differentiability of the loss function, we
use a smoothed form of the $\max$ function, resulting in the final form of the curvature regularization:
%\begin{small}
\begin{equation}
\begin{split}
L_{Curvature}&= {\sum_{i=1}^{C}}-\log(\frac{G_i^{-1}}{\max\{G_1^{-1},\dots,G_C^{-1} \}}) \\
&= \sum_{i=1}^{C}-\log(\frac{G_i^{-1}}{\log(\sum_{i=1}^{C}e^{G_i^{-1}})})
\nonumber
\end{split}
\end{equation}
%\end{small}
As shown in Fig.\ref{fig15}, the perceptual manifold with the smallest curvature produces no loss, and the larger the curvature, the larger the loss. $L_{Curvature}$ causes the curvature of all the perceptual manifolds to converge to the value with the smallest curvature while achieving equilibrium.

In the following, we verify whether $L_{Curvature}$ satisfies the three principles one by one.

\begin{itemize}
\item[\textbf{(1)}] When the curvature $G_i$ of the perceptual manifold is larger, $G_i^{-1}$ is smaller. Since $-\log(\cdot )$ is monotonically decreasing, $-\log(\frac{G_i^{-1}}{\max\{G_1^{-1},\dots,G_C^{-1} \}} )$ increases with $G_i$ increases. $L_{Curvature}$ is consistent with Principle 1.

\item[\textbf{(2)}] When $G_1=\dots=G_C$, $\max\{G_1^{-1},\dots,G_C^{-1}\}=G_1^{-1}=\dots=G_C^{-1}$, so $-\log(\frac{G_i^{-1}}{\max\{G_1^{-1},\dots,G_C^{-1} \}} )=0,i=1,\dots,C$. $L_{Curvature}$ follows Principle 2.

\item[\textbf{(3)}] The curvature penalty term of the perceptual manifold $M^i$ is $0$ when $G_i=\min\{G_1,\dots,G_C\}$. Since the greater the curvature, the greater the penalty, our method aims to bring the curvature of all perceptual manifolds down to $\min\{G_1,\dots,G_C\}$. Obviously, ${\textstyle \sum_{i=1}^{C}}G_i\ge C\cdot  \min\{G_1,\dots,G_C\}$, so our approach promotes curvature balance while also making all perceptual manifolds flatter, which satisfies Principle 3.
\end{itemize}

The curvature regularization can be combined with any loss function. Since the correlation between curvature and accuracy increases with training, we balance the curvature regularization with other losses using a logarithmic function with a hyperparameter $\tau$, and the overall loss is denoted as
\begin{small}
\begin{equation}
\begin{split}
L=L_{original}+\frac{\log_{\tau}{epoch}}{(\frac{L_{Curvature}}{L_{original}}).detach()} \times L_{Curvature},\ \tau>1.
\nonumber
\end{split}
\end{equation}
\end{small}

The term $(\frac{L_{Curvature}}{L_{original}}).detach()$ aims to make the curvature regularization loss of the same magnitude as the original loss. The term \( \log_\tau epoch \) serves to gradually increase the weight of the curvature regularization term in the overall loss as training progresses, thereby amplifying its impact. Specifically, \( \log_\tau \text{epoch}, \tau > 1 \) is an increasing function, where epoch represents the training cycle. Clearly, when \( \tau = epoch \), \( \log_\tau epoch = 1 \), which means that at this point, the curvature regularization term and \( L_{\text{original}} \) have the same influence on the overall loss. Therefore, the hyperparameter \( \tau \) controls when the influence of curvature regularization surpasses \( L_{\text{original}} \). Assuming the total number of training epochs is 200, setting \( \tau \) to 100 indicates that after epoch 100, the impact of curvature regularization will exceed that of \( L_{\text{original}} \). We investigate reasonable values of $\tau$ in experiments (Sec \ref{sec6.2}). The design principle of curvature regularization is compatible with the learning objective of the model, and our experiments show that the effect of curvature imbalance on model bias has been neglected in the past. Thus curvature regularization is not in conflict with $L_{original}$, as evidenced by our outstanding performance on multiple datasets.

\subsection{Dynamic Curvature Regularization (DCR)}
\label{sec5.3}

The curvature of perceptual manifolds varies with the model parameters during training, so it is necessary to update the curvature of each perceptual manifold in real-time. However, there is a challenge: only one batch of features is available at each iteration, and it is not possible to obtain all the features to calculate the curvature of the perceptual manifolds. If the features of all images from the training set are extracted using the current network at each iteration, it will greatly increase the time cost of training.

\begin{algorithm}[h]
\caption{End-to-end training with DCR}
\label{alg1}
%\footnotesize{
\textbf{Require}: Training set $D=\{(x_i,y_i)\}_{i=1}^{M}$. A CNN $\{f(x,\theta_1),g(z,\theta_2)\}$, where $f(\cdot)$ and $g(\cdot)$ denote the feature sub-network and classifier, respectively. The training epoch is $N$.
\begin{algorithmic}[1] %[1] enables line numbers
\STATE Initialize the storage pool Q
\FOR{$epoch = 1$ to $N$}
\FOR{$iteration = 0$ to $\frac{M}{batch \ size} $}
\STATE Sample a mini-batch $\{(x_i,y_i)\}_{i=1}^{batch\ size}$ from $D$.
\STATE Calculate feature embeddings $z_i = f(x_i,\theta_1),i=1,\dots,batch\ size$.
\STATE Store $z_i$ and label $y_i$ into $Q$.
\IF {$epoch< n$}
\IF {$epoch>1$}
\STATE Dequeue the oldest batch of features from $Q$.
\ENDIF 
\STATE Calculate loss $L_{original}$.
\ELSE
\STATE Dequeue the oldest batch of features from $Q$.
\STATE Calculate the curvature of each perceptual manifold.
\STATE Calculate loss: \\ $L=L_{original}+\frac{\log_{\tau}{epoch}}{(\frac{L_{Curvature}}{L_{original}}).detach()} \times L_{Curvature}$.
\ENDIF
\STATE Perform back propagation: $L.backward()$.
\STATE $optimizer.step()$.
\ENDFOR
\ENDFOR
\end{algorithmic}
\end{algorithm}

Inspired by \cite{paper15,paper28}, we design a first-in-first-out storage pool to store the latest historical features of all images. The slow drift phenomenon of features found by \cite{paper54} ensures the reliability of using historical features to approximate the current features. We show the training process in Algorithm \ref{alg1}. Specifically, the features of all batches are stored in the storage pool at the first epoch. To ensure that the drift of the features is small enough, it is necessary to train another $n$ epochs to update the historical features. Experiments of \cite{paper28} on large-scale datasets show that $n$ taken as $5$ is sufficient, so $n$ is set to $5$ in this work. When $epoch>n$, the oldest batch of features in the storage pool is replaced with new features at each iteration, and the curvature of each perceptual manifold is calculated using all features in the storage pool. The curvature regularization term is updated based on the latest curvature.
\textbf{It should be noted} that for decoupled training, CR is applied in the feature learning stage. Our method is employed in training only and does not affect the inference speed of the model.

\section{Experiments}
\label{sec6}

We comprehensively evaluate the effectiveness and generality of curvature regularization on both long-tailed and non-long-tailed datasets. The experiment is divided into two parts, the first part tests curvature regularization on four long-tailed datasets, namely CIFAR-10-LT, CIFAR-100-LT \cite{paper4}, ImageNet-LT \cite{paper4,paper26}, and iNaturalist2018 \cite{paper24}. The second part validates the curvature regularization on two non-long tail datasets, namely CIFAR-100 \cite{paper25} and ImageNet \cite{paper26}. In addition, we train models on CIFAR-100, CIFAR-10/100-LT with a single NVIDIA 2080Ti GPU and ImageNet, ImageNet-LT, and iNaturalist2018 with eight NVIDIA 2080Ti GPUs.

\subsection{Datasets and Evaluation Metrics}
\label{sec6.1}

We conducted experiments on artificially created CIFAR-10-LT, CIFAR-100-LT \cite{paper4}, ImageNet-LT \cite{paper4,paper26}, and real-world long-tailed iNaturalist2018 \cite{paper24} to validate the effectiveness and generalizability of our method. For a fair comparison, the training and test images of all datasets are officially split, and the Top-1 accuracy on the test set is utilized as a performance metric.

\begin{itemize}
  \item \textbf{CIFAR-10-LT} and \textbf{CIFAR-100-LT} are long-tailed datasets including five imbalance factors (IF = $10, 20, 50, 100, 200$) generated based on CIFAR-10 and CIFAR-100, respectively. The imbalance factor (IF) is defined as the value of the number of the most frequent class training samples divided by the number of the least frequent class training samples.

  \item \textbf{ImageNet-LT} is a long-tailed subset of ILSVRC 2012 with an imbalance factor of $256$, which contains $1000$ classes totaling $115.8k$ images, with a maximum of $1280$ images and a minimum of $5$ images per class. The balanced $50$k images were used for testing.

  \item The \textbf{iNaturalist} species classification dataset is a large-scale real-world dataset that suffers from an extremely unbalanced label distribution. The $2018$ version we selected consists of $437,513$ images from $8,142$ classes. The maximum class is $1,000$ images and the minimum class is $2$ images (IF = $500$).
  \item We use the \textbf{ILSVRC2012} split contains $1,281,167$ training and $50,000$ validation images. Each class of \textbf{CIFAR-100} contains $500$ images for training and $100$ images for testing.
\end{itemize}

\subsection{Implementation Details}
\label{sec6.1.2}

\textbf{CIFAR-10/100-LT.} To set up a fair comparison, we used the same random seed to make CIFAR-10/100-LT, and followed the implementation of \cite{paper2}. Consistent with previous studies \cite{paper4,paper28,paper58}, we trained ResNet-32 by SGD optimizer with a momentum of $0.9$, and a weight decay of $2\times 10^{-4}$.

\textbf{ImageNet-LT and iNaturalist2018.} We use ResNext-50 \cite{paper20} on ImageNet-LT and ResNet-50 \cite{paper21} on iNaturalist2018 as the network backbone for all methods. Following previous studies \cite{paper4,paper28,paper58}, we conduct model training with the SGD optimizer based on batch size $256$ (for ImageNet-LT) / $512$ (for iNaturalist), momentum $0.9$, weight decay factor $0.0005$, and learning rate $0.1$ (linear LR decay).

\textbf{ImageNet and CIFAR-100.} Following widely used settings \cite{paper26,paper21,paper23,paper28}, on ImageNet, we use random clipping, mixup \cite{paper22}, and cutmix \cite{paper23} to augment the training data, and all models are optimized by Adam with batch size of $512$, learning rate of $0.05$, momentum of $0.9$, and weight decay factor of $0.0005$. On CIFAR-100, we set the batch size to $64$ and augment the training data using random clipping, mixup, and cutmix. An Adam optimizer with learning rate of $0.1$ (linear decay), momentum of $0.9$, and weight decay factor of $0.005$ is used to train all networks.

\subsection{Effect of $\tau$}
\label{sec6.2}

\begin{figure}[h]
%\vskip -0.2in
\centering
\centerline{\includegraphics[width=1.04\columnwidth]{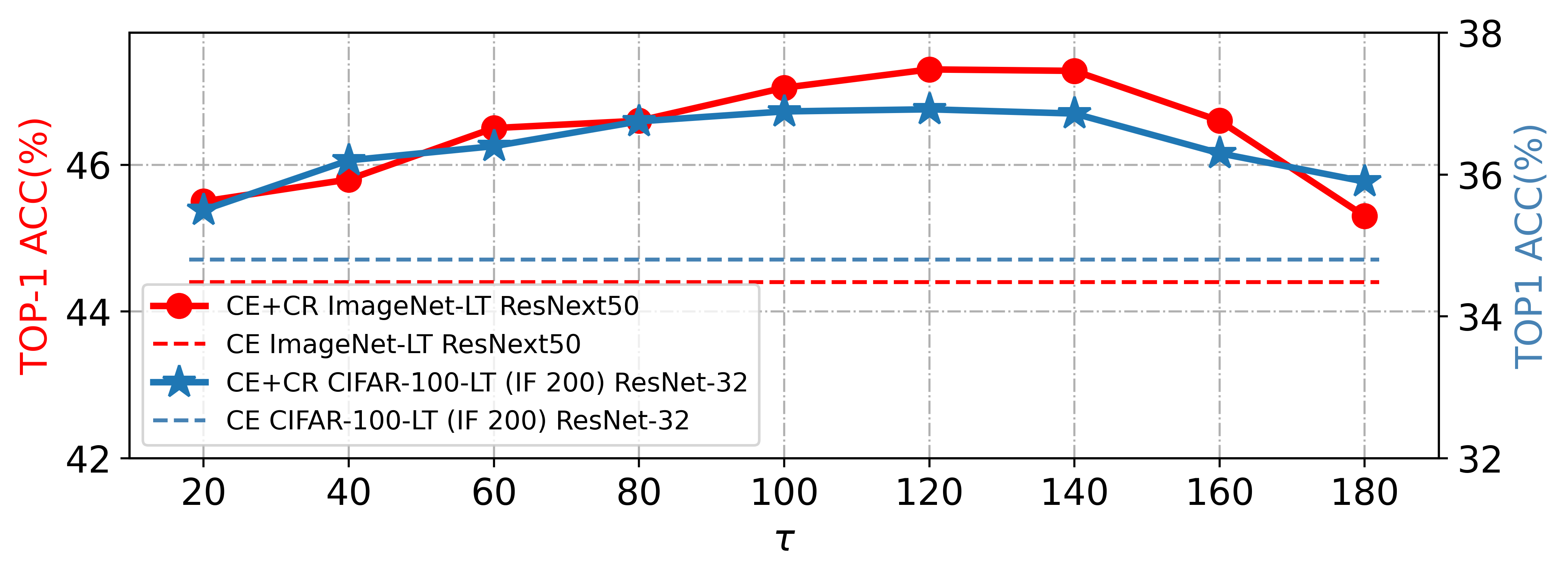}}
\vskip -0.16in
\caption{The effect of $\tau$ on accuracy for both datasets.}
\label{fig7}
%\vskip -0.09in
\end{figure}

When $\tau=epoch$, $\log_{\tau}{epoch}=1$, so the selection of $\tau$ is related to the number of epochs. When the correlation between curvature and accuracy exceeds the correlation between the separation degree and accuracy, we expect $\log_{\tau}{epoch}>1$, which means that the curvature regularization loss is greater than the original loss. Following the \cite{paper8} setting, all models are trained for $200$ epochs, so $\tau$ is less than $200$. To search for the proper value of $\tau$, experiments are conducted for CE + CR with a range of $\tau$, and the results are shown in Fig.\ref{fig7}. Large-scale datasets require more training epochs to keep the perceptual manifolds away from each other, while small-scale datasets can achieve this faster, so we set $\tau=100$ on CIFAR-10/100-LT and CIFAR-100, and $\tau=120$ on ImageNet, ImageNet-LT, and iNaturalist2018.

\subsection{Experiments on Long-Tailed Datasets}
\label{sec6.3}
\subsubsection{Evaluation on CIFAR-10/100-LT}

\begin{table}[t]
%\vskip -0.05in
\caption{Comparison on CIFAR-10-LT. The accuracy (\%) of Top-1 is reported. The best and second-best results are shown in \underline{\textbf{underlined bold}} and \textbf{bold}, respectively.}
\label{table1}
\vskip -0.07in
\centering  
\begin{small}
\renewcommand\arraystretch{1.05}
\setlength{\tabcolsep}{12pt} %修改边距
\begin{tabular}{l|cccc}
\hline \toprule 
Dataset   &\multicolumn{4}{c}{CIFAR-10-LT}     \\ \hline
Backbone Net  &\multicolumn{4}{c}{ResNet-32} \\ \hline
imbalance factor   &200   &100   &50   &10   \\ \hline
MiSLAS \cite{paper1}  &77.3  &82.1 &\textbf{85.7} &\underline{\textbf{90.0}}    \\ 
LDAM-DRW \cite{paper2}  & \multicolumn{1}{c}{-}  &77.0 &81.0 &88.2    \\ \hline

Cross Entropy   & 65.6  &70.3  &74.8  &86.3      \\ 
+ CR     & 67.9  &72.6  &76.2  &89.5     \\  \hline

Focal Loss \cite{paper3} & 65.2  &70.3  &76.7  &86.6      \\ 
+ CR   & 67.3  &71.8  &79.1  &88.4     \\  \hline

CB Loss \cite{paper4} & 68.8  &74.5  &79.2  &87.4     \\ 
+ CR     & 70.3  &75.8  &79.8  &89.1      \\  \hline

BBN \cite{paper5}     & \multicolumn{1}{c}{-}   & 79.8  &82.1   &88.3    \\ 
+ CR \cite{paper9}     & \multicolumn{1}{c}{-}   & 81.2  &83.5   &89.4   \\  \hline

De-c-TDE \cite{paper6} & \multicolumn{1}{c}{-}  &80.6  &83.6  &88.5     \\ 
+ CR    &\multicolumn{1}{c}{-}  &81.8  &84.5  &\textbf{89.9}      \\  \hline

GCL \cite{paper9}    &\textbf{79.0}  & \textbf{82.7}  &85.5  &\multicolumn{1}{c}{-}       \\
+ CR    &\underline{\textbf{79.9}}  & \underline{\textbf{83.5}}  &\underline{\textbf{86.8}}  &\multicolumn{1}{c}{-}   \\

\bottomrule \hline
\end{tabular}
\end{small}
%\vskip -0.18in
\end{table}

\begin{table}[t]
%\vskip -0.05in
\caption{Comparison on CIFAR-100-LT. The accuracy (\%) of Top-1 is reported. The best and second-best results are shown in \underline{\textbf{underlined bold}} and \textbf{bold}, respectively.}
\label{table1_2}
\vskip -0.07in
\centering  
\begin{small}
\renewcommand\arraystretch{1.15}
\setlength{\tabcolsep}{12pt} %修改边距
\begin{tabular}{l|cccc}
\hline \toprule 
Dataset     &\multicolumn{4}{c}{CIFAR-100-LT}  \\ \hline
Backbone Net  &\multicolumn{4}{c}{ResNet-32} \\ \hline
imbalance factor    &200   &100   &50   &10 \\ \hline
MiSLAS \cite{paper1}   &42.3 &47.0 &52.3 & \underline{\textbf{63.2}}    \\ 
LDAM-DRW \cite{paper2}   & \multicolumn{1}{c}{-} &42.0 &46.6 & 58.7   \\ \hline

Cross Entropy     &34.8  & 38.2   & 43.8   &55.7   \\ 
+ CR       &36.9  & 40.5   & 45.1   &57.4  \\  \hline

Focal Loss \cite{paper3}    &35.6  & 38.4   & 44.3   &55.7   \\ 
+ CR      &37.5  & 40.2   & 45.2   &58.3   \\  \hline

CB Loss \cite{paper4}    &36.2  & 39.6   & 45.3   &57.9   \\ 
+ CR       &38.5  & 40.7   & 46.8   &59.2   \\  \hline

BBN \cite{paper5}      & \multicolumn{1}{c}{-}   &42.5  &47.0  &59.1 \\ 
+ CR \cite{paper9}        & \multicolumn{1}{c}{-}   &43.7 &48.1  &60.0 \\  \hline

De-c-TDE \cite{paper6}    &\multicolumn{1}{c}{-}  & 44.1   & 50.3   &59.6   \\ 
+ CR      &\multicolumn{1}{c}{-}  & 45.7   & 51.4   &60.3   \\  \hline

RIDE (4*) \cite{paper7}    & \multicolumn{1}{c}{-}   &  48.7  & \textbf{59.0}  & 58.4   \\ 
+ CR   & \multicolumn{1}{c}{-}  & 49.8   & \underline{\textbf{59.8}}   &59.5   \\  \hline

RIDE + CMO \cite{paper8}    &\multicolumn{1}{c}{-} &\textbf{50.0} &53.0 &60.2   \\  
+ CR  &\multicolumn{1}{c}{-} &\underline{\textbf{50.7}} &54.3 &\textbf{61.4}  \\ \hline

GCL \cite{paper9}      &\textbf{44.9}  &48.7  &53.6  &\multicolumn{1}{c}{-}     \\
+ CR      &\underline{\textbf{45.6}}  &49.8  &55.1  &\multicolumn{1}{c}{-}  \\

\bottomrule \hline
\end{tabular}
\end{small}
\vskip -0.05in
\end{table}

Tables \ref{table1} and \ref{table1_2} summarizes the improvements of CR for several state-of-the-art methods on long-tailed CIFAR-10 and CIFAR-100, and we observe that CR significantly improves all methods. For example, in the setting of IF $200$, CR results in performance gains of $2.3\%$, $2.1\%$, and $1.5\%$ for CE, Focal loss \cite{paper3}, and CB loss \cite{paper4}, respectively. When CR is applied to feature training, the performance of BBN \cite{paper5} is improved by more than $1\%$ on each dataset, which again validates that curvature imbalance negatively affects the learning of classifiers. When CR is applied to several state-of-the-art methods (e.g., RIDE + CMO \cite{paper8} (2022) and GCL \cite{paper9} (2022)), CR achieved higher classification accuracy with all IF settings of CIFAR-100-LT (Table \ref{table1_2}).

\begin{table}[!t]
%\vskip -0.25in
\caption{Top-1 accuracy (\%) of ResNext-50 \cite{paper20} on ImageNet-LT for classification. The best and the second-best results are shown in \underline{\textbf{underline bold}} and \textbf{bold}, respectively. }
%\caption{Top-1 accuracy (\%) of ResNext-50 \cite{paper20} on ImageNet-LT and Top-1 accuracy (\%) of ResNet-50 \cite{paper21} on iNaturalist2018 for classification. The best and the second-best results are shown in \underline{\textbf{underline bold}} and \textbf{bold}, respectively. }
\label{table2}
\vskip -0.07in
\centering  
\begin{small}
\renewcommand\arraystretch{1.15}
\setlength{\tabcolsep}{8pt} %修改边距
\begin{tabular}{l|cccc}
\hline \toprule
\multirow{3}{*}{Methods}    & \multicolumn{4}{c}{ImageNet-LT}    \\ \cline{2-5}
& \multicolumn{4}{c}{ResNext-50}   \\ \cline{2-5}
& Head   &Middle  &Tail   &Overall    \\ \hline
OFA \cite{paper10}    &47.3 & 31.6 & 14.7 & 35.2  \\
DisAlign \cite{paper11}  &59.9 &49.9 &31.8 &52.9   \\
MiSLAS \cite{paper1}   &65.3 &50.6 & 33.0 & 53.4  \\ 
DiVE \cite{paper12} &64.0 & 50.4 & 31.4 & 53.1  \\  
PaCo \cite{paper13} &63.2 & 51.6 & 39.2 & 54.4  \\ 
GCL \cite{paper9} &\multicolumn{1}{c}{-}  & \multicolumn{1}{c}{-} & \multicolumn{1}{c}{-} & 54.9   \\ \hline

CE &65.9 & 37.5 & 7.70 & 44.4 \\
+ CR &65.1 & 40.7 & 19.5 & 47.3  \\ \hline

Focal Loss \cite{paper3} &67.0 & 41.0 & 13.1 & 47.2  \\
+ CR  &67.3 & 43.2 & 22.5 &49.6  \\ \hline

BBN \cite{paper5}   &43.3 & 45.9 & \textbf{43.7}  & 44.7  \\  
+ CR & 45.2& 46.8 & \underline{\textbf{44.5}} & 46.2  \\  \hline

LDAM \cite{paper2} &60.0 & 49.2 & 31.9 & 51.1  \\
+ CR &60.8 & 50.3 &33.6 &52.4  \\ \hline

LADE \cite{paper14} &62.3 & 49.3 & 31.2 & 51.9  \\
+ CR &62.5 & 50.1 &33.7 &53.0  \\ \hline

MBJ \cite{paper15} &61.6 & 48.4 & 39.0 & 52.1  \\
+ CR &62.8 & 49.2 & 40.4 &53.4  \\ \hline

RIDE (4*) \cite{paper7}  &\underline{\textbf{67.8}} &53.4 &36.2 &56.6   \\ 
+ CR  &\underline{\textbf{68.5}} &54.2 &38.8 &\underline{\textbf{57.8}}   \\ \hline

%GCL \cite{paper43}  &CVPR & \multicolumn{1}{c}{-} &\multicolumn{1}{c}{-}  &\multicolumn{1}{c}{-}  &\textbf{54.9} &\multicolumn{1}{c}{-}  &\multicolumn{1}{c}{-}  &\multicolumn{1}{c}{-}  &\textbf{72.0}  \\
RIDE + CMO \cite{paper8}  &66.4 & \underline{\textbf{54.9}} & 35.8 & 56.2  \\ 
+ CR      &67.3 & \textbf{54.6} &38.4 & \textbf{57.4}  \\
\bottomrule \hline
 \end{tabular}
 \end{small}
% \vskip -0.1in
 \end{table}

\begin{table}[!t]
%\vskip -0.25in
\caption{Top-1 accuracy (\%) of ResNet-50 \cite{paper21} on iNaturalist2018 for classification. The best and the second-best results are shown in \underline{\textbf{underline bold}} and \textbf{bold}, respectively. }
%\caption{Top-1 accuracy (\%) of ResNext-50 \cite{paper20} on ImageNet-LT and Top-1 accuracy (\%) of ResNet-50 \cite{paper21} on iNaturalist2018 for classification. The best and the second-best results are shown in \underline{\textbf{underline bold}} and \textbf{bold}, respectively. }
\label{table2_2}
\vskip -0.07in
\centering  
\begin{small}
\renewcommand\arraystretch{1.1}
\setlength{\tabcolsep}{8pt} %修改边距
\begin{tabular}{l|cccc}
\hline \toprule
\multirow{3}{*}{Methods}    & \multicolumn{4}{c}{iNaturalist 2018}  \\ \cline{2-5}
 & \multicolumn{4}{c}{ResNet-50}  \\ \cline{2-5}
&Head   &Middle   &Tail   &Overall \\ \hline
OFA \cite{paper10}     & \multicolumn{1}{c}{-}  & \multicolumn{1}{c}{-} & \multicolumn{1}{c}{-} & 65.9 \\
DisAlign \cite{paper11}   &68.0 &71.3 &69.4 &70.2  \\
MiSLAS \cite{paper1}    &\underline{\textbf{73.2}}  &72.4 & 70.4 & 71.6 \\ 
DiVE \cite{paper12}  & 70.6  & 70.0 & 67.5 & 69.1 \\  
PaCo \cite{paper13}  & 69.5  & 72.3 & 73.1 & 72.3 \\ 
GCL \cite{paper9}  &\multicolumn{1}{c}{-}  & \multicolumn{1}{c}{-} & \multicolumn{1}{c}{-} &72.0  \\ \hline

CE  & 67.2  & 63.0 & 56.2 & 61.7 \\
+ CR  & 67.3  & 62.6 & 61.7 & 63.4 \\ \hline

Focal Loss \cite{paper3}  & \multicolumn{1}{c}{-}  & \multicolumn{1}{c}{-} & \multicolumn{1}{c}{-} & 61.1 \\
+ CR   &69.4  &61.7 &57.2 &62.3 \\ \hline

BBN \cite{paper5}    & 49.4  & 70.8 & 65.3 & 66.3 \\  
+ CR  &50.6 & 71.5 &66.8 & 67.6\\  \hline

LDAM \cite{paper2}  & \multicolumn{1}{c}{-}  & \multicolumn{1}{c}{-} & \multicolumn{1}{c}{-} & 64.6\\
+ CR  & 69.3  &66.7  &61.9 &65.7 \\ \hline

LADE \cite{paper14}  & \multicolumn{1}{c}{-}  & \multicolumn{1}{c}{-} & \multicolumn{1}{c}{-} & 69.7\\
+ CR  &72.5  &70.4  &65.7  &70.6 \\ \hline

MBJ \cite{paper15}  & \multicolumn{1}{c}{-}  & \multicolumn{1}{c}{-} & \multicolumn{1}{c}{-} & 70.0 \\
+ CR  &\textbf{73.1}  &70.3 &66.0 & 70.8 \\ \hline

RIDE (4*) \cite{paper7}   &70.9 &72.4 &73.1 &72.6 \\ 
+ CR   &71.0 & \underline{\textbf{73.8}} &\textbf{74.3} &\textbf{73.5} \\ \hline

%GCL \cite{paper43}  &CVPR & \multicolumn{1}{c}{-} &\multicolumn{1}{c}{-}  &\multicolumn{1}{c}{-}  &\textbf{54.9} &\multicolumn{1}{c}{-}  &\multicolumn{1}{c}{-}  &\multicolumn{1}{c}{-}  &\textbf{72.0}  \\
RIDE + CMO \cite{paper8}   &70.7  & 72.6 &73.4 & 72.8 \\ 
+ CR      &71.6  &\textbf{73.7} &\underline{\textbf{74.9}} &\underline{\textbf{73.8}} \\
\bottomrule \hline
 \end{tabular}
 \end{small}
% \vskip -0.05in
 \end{table}

\subsubsection{Evaluation on ImageNet-LT and iNaturalist2018}

The results on ImageNet-LT and iNaturalist2018 are shown in Tables \ref{table2} and \ref{table2_2}. We not only report the overall performance of CR, but also additionally add the performance on three subsets of Head (more than 100 images), Middle (20-100 images), and Tail (less than 20 images). From Tables \ref{table2} and \ref{table2_2}, we observe the following three conclusions: first, CR results in significant overall performance improvements for all methods, including $2.9\%$ and $2.4\%$ improvements on ImageNet-LT for CE and Focal loss, respectively. Second, when CR is combined with feature training, the overall performance of BBN \cite{paper5} is improved by $1.5\%$ and $1.3\%$ on the two datasets, respectively, indicating that curvature-balanced feature learning facilitates classifier learning. Third, our approach still boosts model performance when combined with advanced techniques (RIDE \cite{paper7} (2021), RIDE + CMO \cite{paper8} (2022)), suggesting that curvature-balanced feature learning has not yet been considered by other methods.

\begin{table}[t]
\small
\renewcommand\arraystretch{1.08}
%\vskip -0.14
\setlength{\tabcolsep}{3pt} %修改边距
\caption{Comparison on ImageNet and CIFAR-100.}
%\caption{Comparison on ImageNet and CIFAR-100. On ImageNet, we use random clipping, mixup \cite{paper22}, and cutmix \cite{paper23} to augment the training data, and all models are optimized by Adam with batch size of 512, learning rate of 0.05, momentum of 0.9, and weight decay factor of 0.0005. On CIFAR-100, we set the batch size to 64 and augment the training data using random clipping, mixup, and cutmix. An Adam optimizer with learning rate of 0.1 (linear decay), momentum of 0.9, and weight decay factor of 0.005 is used to train all networks.}
\vskip -0.07in
\label{table3}
\centering  
\begin{small}
\begin{tabular}{l|ccc|ccc}
\hline  \toprule
   & \multicolumn{3}{c|}{ImageNet}  &  \multicolumn{3}{c}{CIFAR-100}  \\ \hline
Methods & CE  & CE + CR & $\Delta$ & CE & CE + CR & $\Delta$ \\ \hline
VGG16 \cite{paper16}&  71.6 & 72.7 &+1.1 & 71.9 & 73.2 &+1.3 \\
%VGG19 &  72.1 & 73.3 & +1.2 & 71.2 & 72.5 & +1.3 \\
BN-Inception \cite{paper17} &  73.5 & 74.3 &+0.8 & 74.1 & 75.0 &+0.9 \\
ResNet-18 \cite{paper21} &  70.1 & 71.3 &+1.2  &75.6   & 77.1 &+1.5  \\
ResNet-34 \cite{paper21} &  73.5 & 74.6 &+1.1  & 76.8 & 78.0 &+1.2  \\
ResNet-50 \cite{paper21} &  76.0 & 76.8 &+0.8 & 77.4 & 78.3 &+0.9  \\
DenseNet-201 \cite{paper18} &  77.2 & 78.0 &+0.8  & 78.5 & 79.7 &+1.2  \\
SE-ResNet-50 \cite{paper19} &  77.6 & 78.3 &+0.7  & 78.6 & 79.5 &+0.9  \\
ResNeXt-101 \cite{paper20} &  78.8 & 79.7 &+0.9  & 77.8  & 78.9 & +1.1  \\
 \bottomrule \hline
 \end{tabular}
\end{small}
%\vskip -0.14in
\end{table}

\subsection{Experiments on Non-Long-Tailed Datasets}
\label{sec6.4}
%\subsubsection{Evaluation on CIFAR-100 and ImageNet}

Curvature imbalance may still exist on sample-balanced datasets, so we evaluate CR on non-long-tailed datasets. Table \ref{table3} summarizes the improvements of CR on CIFAR-100 and ImageNet for various backbone networks, and we observe that CR results in approximately $1\%$ performance improvement for all backbone networks. In particular, the accuracy of CE + CR exceeds CE by $1.5\%$ on CIFAR-100 when using ResNet-18 \cite{paper21} as the backbone network. The experimental results show that our proposed curvature regularization is applicable to non-long-tailed datasets and compatible with existing backbone networks and methods.

\subsection{Curvature Regularization Reduces Model Bias}

\begin{figure}[t]
%\vskip -0.05in
\centering
\centerline{\includegraphics[width=1\columnwidth]{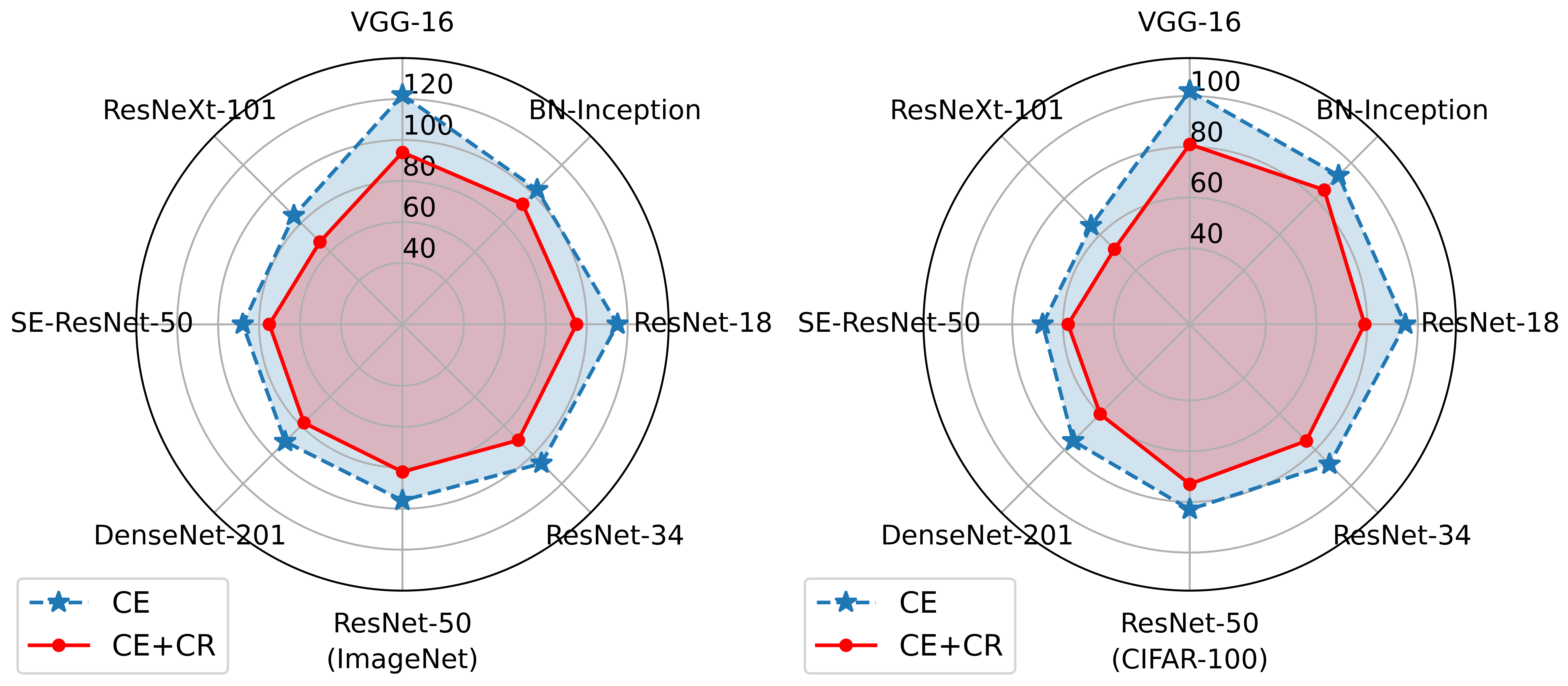}}
%\vskip -0.05in
\caption{Curvature regularization reduces the bias of multiple backbone networks trained on ImageNet and CIFAR-100.}
\label{fig8}
\vskip -0.1in
\end{figure}

Here we explore how curvature regularization improves the model performance. Measuring the model bias with the variance of the accuracy of all classes \cite{paper27}.
Fig.\ref{fig1} presents a comparison of the model bias before and after applying our proposed curvature regularization to existing long-tailed recognition methods (GCL, BBN, Focal Loss, CB Loss, and CE Loss) on long-tailed datasets CIFAR-100-LT and ImageNet-LT. Fig.\ref{fig8} demonstrates the model bias on relatively balanced datasets CIFAR-100 and ImageNet, using multiple backbone networks, comparing cross-entropy loss (CE) with and without our proposed curvature regularization. The results show that curvature regularization consistently reduces model bias.

\begin{figure}[t]
\centering
	\begin{minipage}{0.495\linewidth}
		\centering
		\includegraphics[width=1\linewidth]{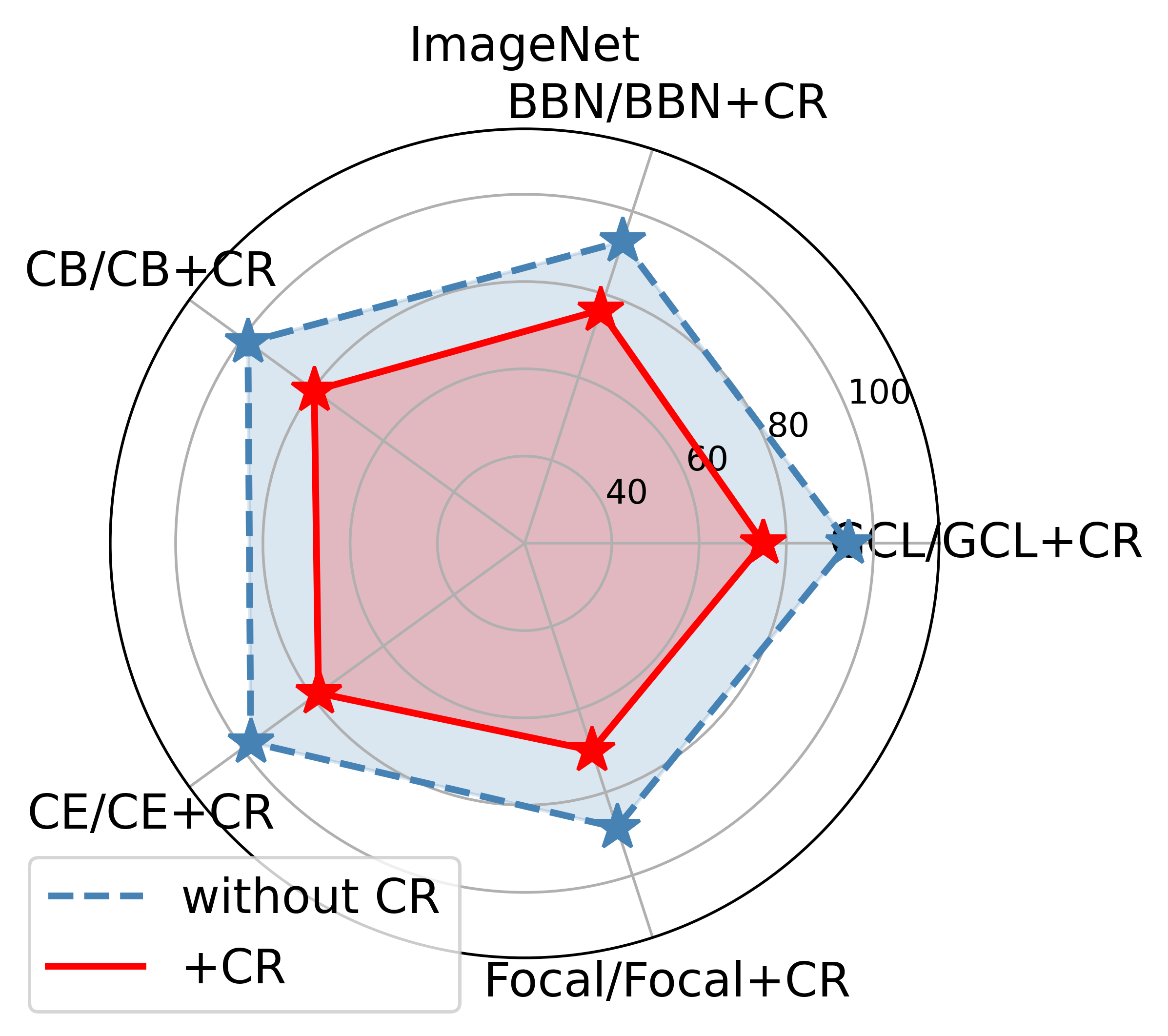}
		%\caption{chutian1}
		%\label{chutian1}%文中引用该图片代号
	\end{minipage}
	\begin{minipage}{0.495\linewidth}
		\centering
		\includegraphics[width=1\linewidth]{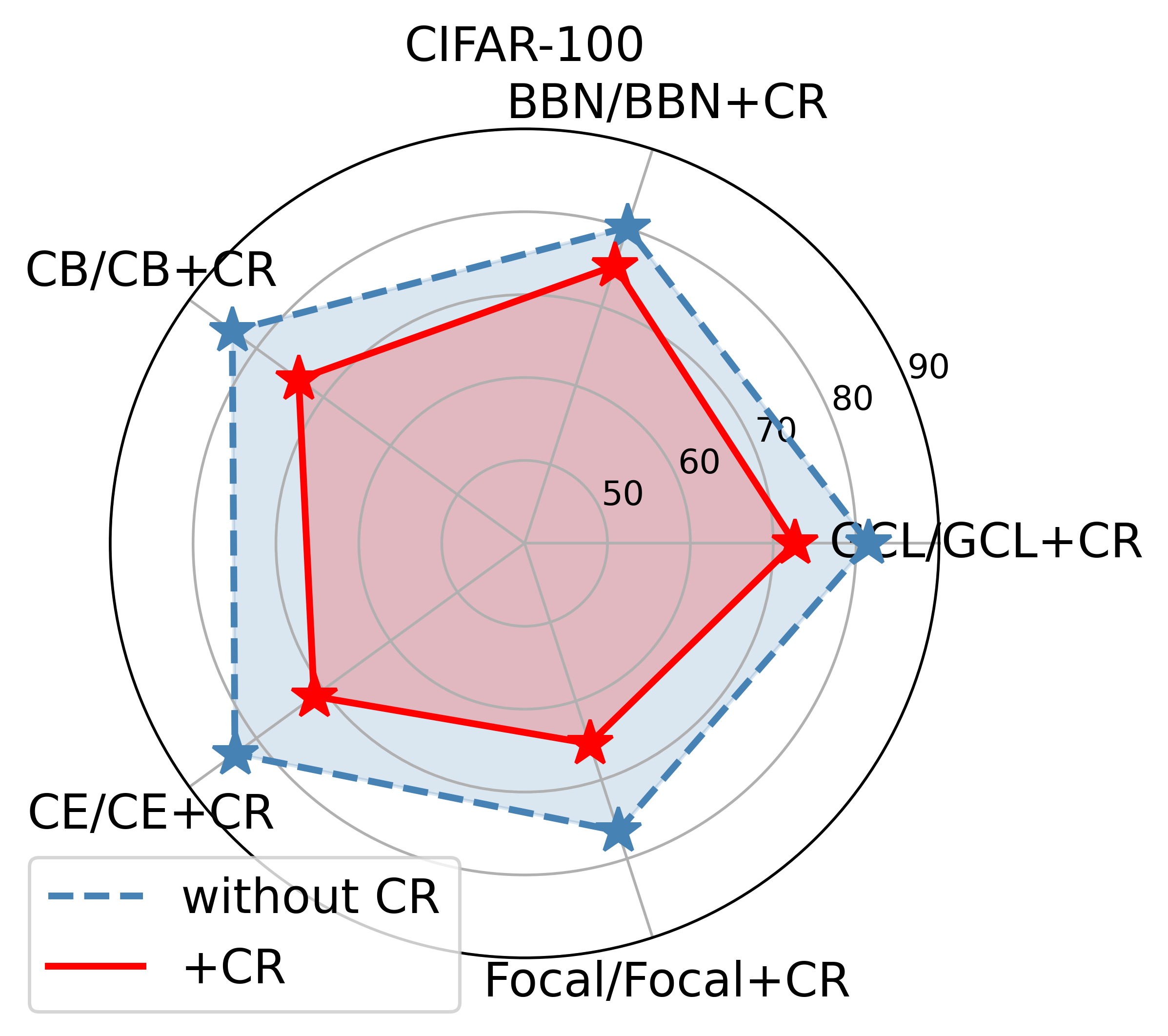}
		%\caption{chutian2}
		%\label{chutian2}%文中引用该图片代号
	\end{minipage}
\caption{Comparison of model bias before and after applying curvature regularization to various long-tailed recognition methods on relatively balanced datasets (ImageNet and CIFAR-100).}
\label{fig9}
%\vskip -0.2in
\end{figure}

Furthermore, we employed ResNet-50 as the backbone network on the perfectly balanced CIFAR-100 and the relatively balanced ImageNet datasets. We initially trained models using existing long-tailed recognition methods such as GCL, BBN, CB Loss, and Focal Loss, followed by retraining the models with added curvature regularization on top of these methods. The experimental results, as shown in Fig.\ref{fig9}, reveal that without curvature regularization, the bias of models trained with GCL, BBN, CB Loss, and Focal Loss was almost identical to or only slightly lower than that of models trained with standard CE Loss. This supports our viewpoint that long-tailed recognition methods designed based on sample numbers are less effective in non-long-tailed scenarios. However, after applying curvature regularization, the model bias was significantly reduced, further validating the generality and effectiveness of curvature regularization.
By combining Tables \ref{table1} and \ref{table2}, it can be found that curvature regularization reduces the model bias mainly by improving the performance of the tail class and does not compromise the performance of the head class, thus improving the overall performance.

\subsection{Curvature Regularization Promotes Convergence}

\begin{figure}[h]
\centering
	\begin{minipage}{0.495\linewidth}
		\centering
		\includegraphics[width=1\linewidth]{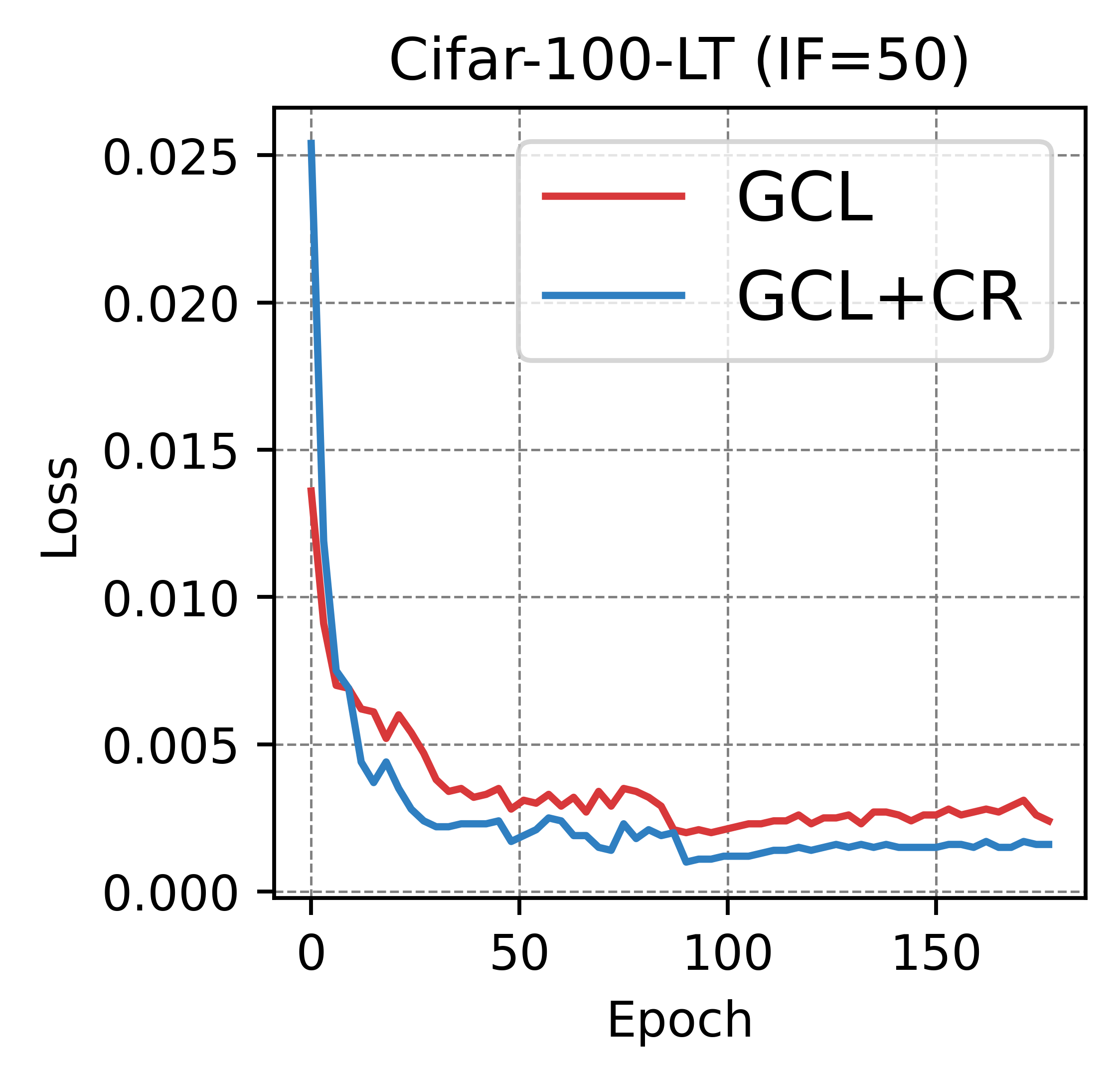}
		%\caption{chutian1}
		%\label{chutian1}%文中引用该图片代号
	\end{minipage}
	\begin{minipage}{0.495\linewidth}
		\centering
		\includegraphics[width=1\linewidth]{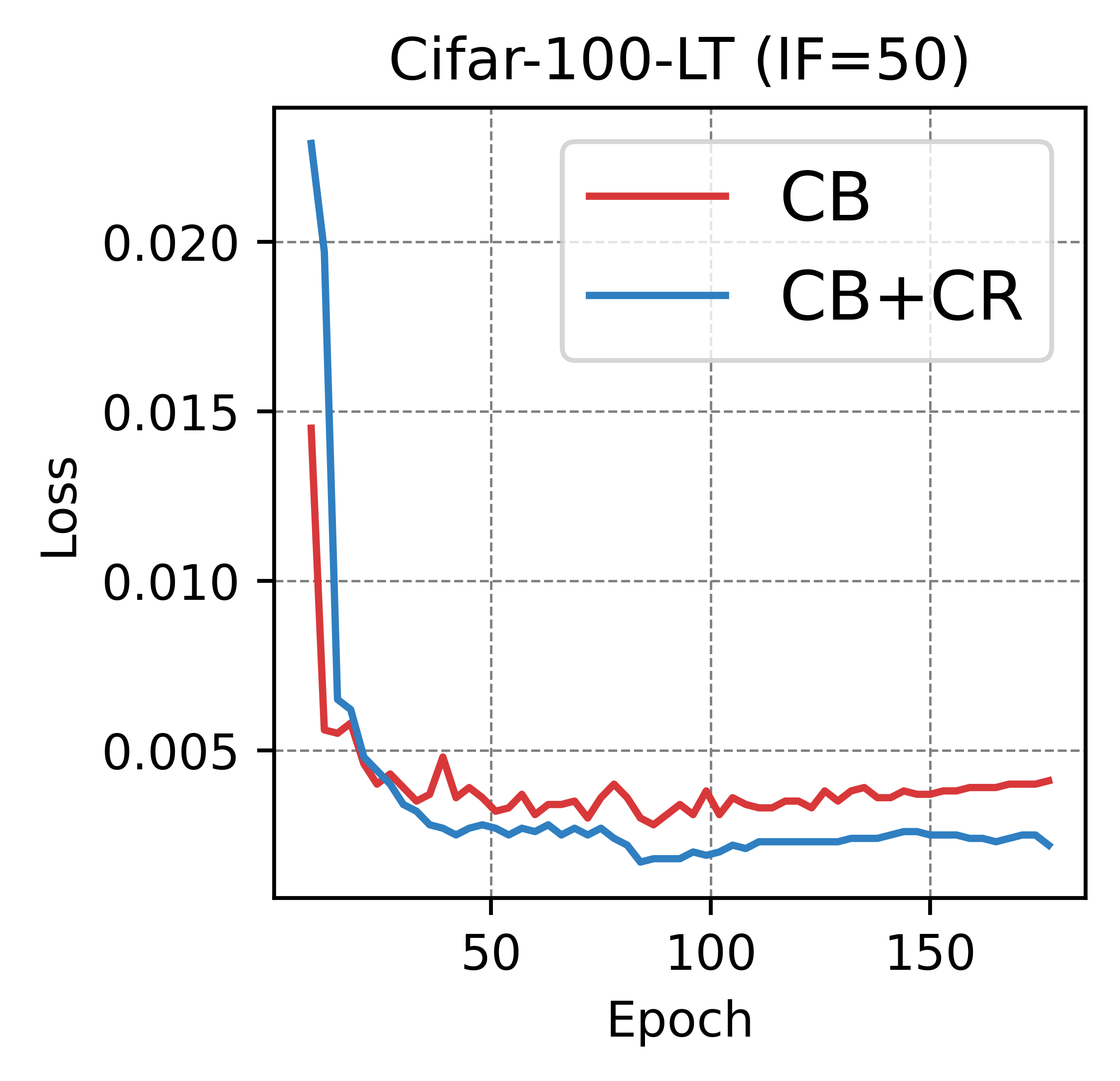}
		%\caption{chutian2}
		%\label{chutian2}%文中引用该图片代号
	\end{minipage}
\caption{Classification loss curves with and without CR.}
\label{fig10}
%\vskip -0.2in
\end{figure}

In Fig.\ref{fig10}, we present the classification loss curves of the long-tailed recognition methods GCL and CB Loss on the CIFAR-100 dataset with an imbalance factor of 50, both before and after applying CR. It is important to note that GCL, as a method for enhancing image embeddings, still adopts cross-entropy as its classification loss. It can be observed that CR facilitates faster convergence of the classification loss and achieves a lower final loss value. This observation supports our hypothesis that flatter and simpler perceptual manifolds are more conducive to classification, thereby accelerating the convergence of the classification loss.

\subsection{More Analysis of Curvature Regularization}

Here, we explored the following two questions:
\begin{itemize}
\item[(1)] Is the curvature more balanced after training with CR? 
\item[(2)] Did the correlation between curvature imbalance and class accuracy decrease after training with CR?
\end{itemize}

Recall that in Section \ref{sec6.4}, we trained multiple backbone networks on ImageNet and CIFAR-100. The features of all samples were extracted using ResNet-18 and VGG-16 which was trained on ImageNet and CIFAR-100 with CE and with CE + CR, respectively, and the curvature of each perceptual manifold was calculated. The degree of imbalance is measured by the variance of the curvature of all perceptual manifolds; the larger the variance, the more imbalanced the curvature. The experimental results are shown in Table \ref{table4}, where the curvature of the perceptual manifolds represented by the ResNet-18 trained with curvature regularization is more balanced.

\begin{table}[!h]
\centering 
\renewcommand\arraystretch{1.1}
\vskip -0.15in
\setlength{\tabcolsep}{23.5pt} %修改边距
\caption{The variance of the curvature of all perceptual manifolds.}
\vskip -0.06in
\label{table4}
\begin{tabular}{l|cc}
\hline  \toprule
                              & \multicolumn{1}{c|}{ImageNet} & CIFAR-100 \\ \hline
                              & \multicolumn{2}{c}{ResNet-18}            \\ \hline
\multicolumn{1}{c|}{CE}      & \multicolumn{1}{c|}{25.7}     & 20.4      \\ \hline
\multicolumn{1}{c|}{CE + CR} & \multicolumn{1}{c|}{14.2 \textbf{(-11.5)}}     & 11.8 \textbf{(-8.6)}      \\  \hline

                              & \multicolumn{2}{c}{VGG-16}            \\ \hline
\multicolumn{1}{c|}{CE}      & \multicolumn{1}{c|}{27.4}     & 23.5      \\ \hline
\multicolumn{1}{c|}{CE + CR} & \multicolumn{1}{c|}{13.8 \textbf{(-13.6)}}     & 13.3 \textbf{(-10.2)}      \\ 
\bottomrule \hline
\end{tabular}
%\vskip -0.2in
\end{table}

\begin{table}[h]
\centering 
\renewcommand\arraystretch{1.1}
%\vskip -0.14in
\setlength{\tabcolsep}{19pt} %修改边距
\caption{The Pearson correlation coefficient between the curvature of the perceptual manifold and the corresponding class accuracy.}
\vskip -0.07in
\label{table5}
\begin{tabular}{l|cc}
\hline  \toprule
                              & \multicolumn{1}{c|}{ImageNet} & CIFAR-100 \\ \hline
                              & \multicolumn{2}{c}{ResNet-18}            \\ \hline
\multicolumn{1}{c|}{CE}      & \multicolumn{1}{c|}{-0.583}     & -0.648      \\ \hline
\multicolumn{1}{c|}{CE + CR} & \multicolumn{1}{c|}{-0.257 \textbf{(+0.326)}}    & -0.285 \textbf{(+0.363)}     \\  \hline

                              & \multicolumn{2}{c}{VGG-16}            \\ \hline
\multicolumn{1}{c|}{CE}      & \multicolumn{1}{c|}{-0.569}     & -0.635      \\ \hline
\multicolumn{1}{c|}{CE + CR} & \multicolumn{1}{c|}{-0.226 \textbf{(+0.343)} }    & -0.251 \textbf{(+0.384)}     \\ 
\bottomrule \hline
\end{tabular}
%\vskip -0.05in
\end{table}

We still use CE and CE + CR to train ResNet-18 on ImageNet and CIFAR-100, respectively, and then test the accuracy of two ResNet-18 on each class. The features of all samples were extracted using two ResNet-18 and the mean Gaussian curvature of each perceptual manifold was calculated. We calculated the Pearson correlation coefficients between the class accuracy and the curvature of the corresponding perceptual manifold for ResNet-18 trained with CE and with CE + CR, respectively. For VGG-16, the same experiments as for ResNet-18 were performed. The experimental results are presented in Table \ref{table5}, where it can be seen that the negative correlation between the mean Gaussian curvature of the perceptual manifold and the class accuracy decreases significantly after using curvature regularization.

Reflecting on the design principles of Curvature Regularization presented in Section \ref{sec5.1}, we intended for the CR term to impose stronger penalties on perceptual manifolds with higher curvature, driving the curvature of all manifolds toward balance and flatness. Table 6 illustrates that after applying CR, the curvature of perceptual manifolds becomes more balanced, confirming that our proposed CR term meets its intended goals.

Why then does the negative correlation between class accuracy and perceptual manifold curvature diminish? This is because after applying CR, the curvature of the perceptual manifolds becomes more balanced. However, as shown in Figs \ref{fig1}, \ref{fig8}, and \ref{fig9}, while CR significantly reduces model bias, some degree of bias persists. This suggests that multiple factors influence model bias, with perceptual manifold curvature being just one of them. From the geometric perspective we have developed for analyzing model fairness, other geometric properties of perceptual manifolds—such as intrinsic dimensionality and topological complexity—are also potential factors affecting model fairness.

\section{Conclusion and Discussion}
This work mines and explains the impact of data on the model bias from a geometric perspective, introducing the imbalance problem to non-long-tailed data and providing a geometric analysis perspective to drive toward fairer AI.

In the field of object detection, it is often encountered that although a class does not appear frequently, the model can always detect such instances efficiently. It is easy to observe that classes with simple patterns are usually easier to learn, even if the frequency of such classes is low. Therefore, classes with low frequency in object detection are not necessarily always harder to learn. We believe that it is a valuable research direction to analyze the richness of the instances contained in each class, and then pay more attention to the hard classes. The dimensionality of all images or feature embeddings in the image classification task is the same, which facilitates the application of the semantic scale proposed in this paper. However, the non-fixed dimensionality of each instance in the field of object detection brings new challenges, so we have to consider the effect of dimensionality on the semantic scale, which is a direction worthy of further study.

\bibliographystyle{IEEEtran}
%\normalem
\bibliography{TKDE2023}

\begin{IEEEbiography}[{\includegraphics[width=1in,height=1.25in,clip,keepaspectratio]{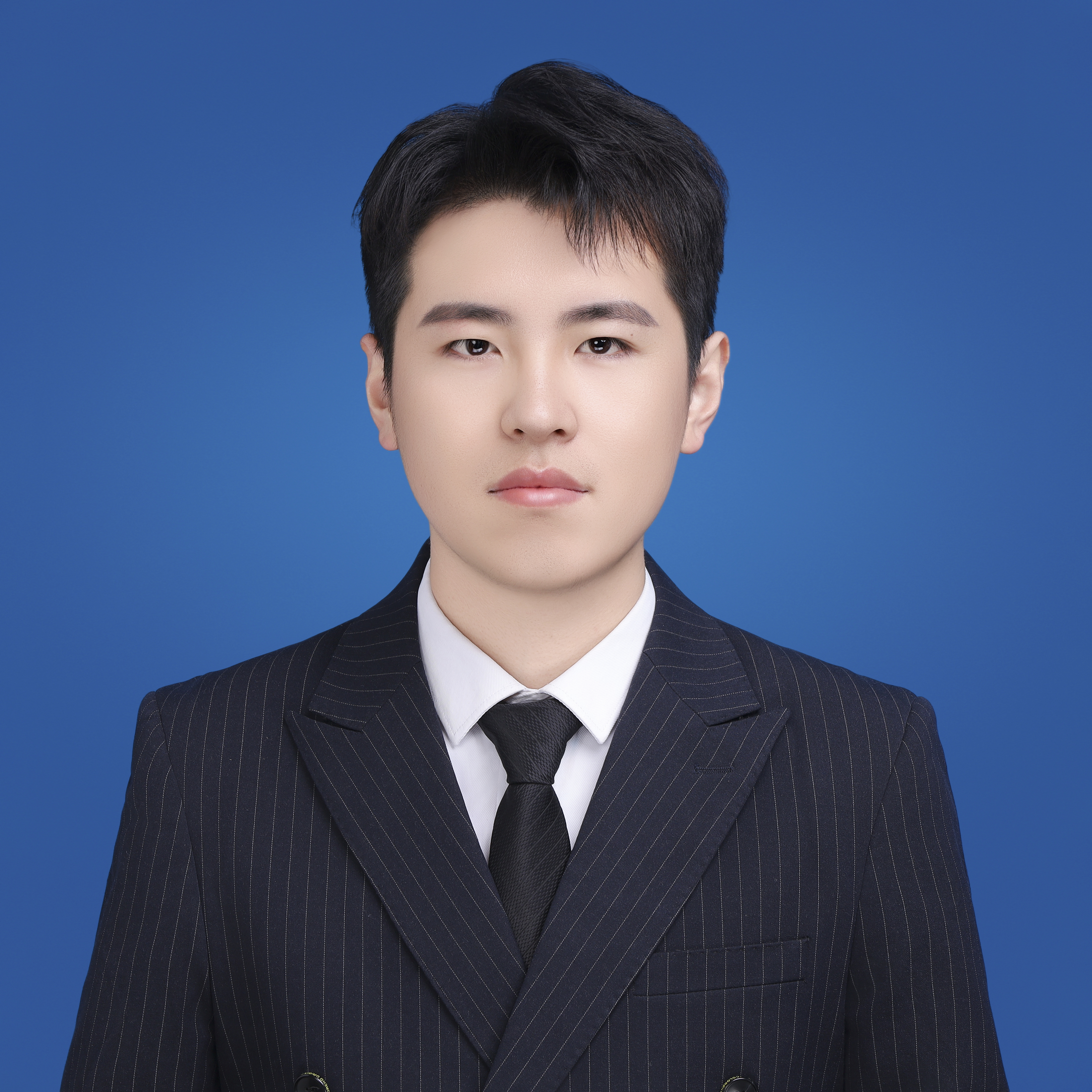}}]{Yanbiao Ma}
received the B.S. degree in Intelligent Science and Technology from Xidian University, China, in 2020.
He is currently pursuing the Ph.D. degree with the Key Laboratory of Intelligent Perception and Image Understanding of the Ministry of Education, School of Artificial Intelligence Xidian University, China. He has been dedicated to research in the field of imbalanced learning and fairness in deep neural networks.
\end{IEEEbiography}

%\vspace{-1cm}

\begin{IEEEbiography}[{\includegraphics[width=1in,height=1.25in,clip,keepaspectratio]{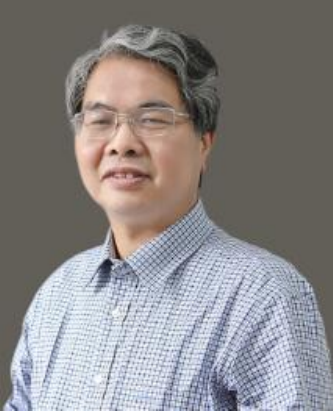}}]{Licheng Jiao} (Fellow, IEEE) received the B.S. degree from Shanghai Jiaotong University, Shanghai, China, in 1982 and the M.S. and PhD degree from Xi’an Jiaotong University, Xi’an, China, in 1984 and 1990, respectively.

Since 1992, he has been a distinguished professor with the school of Electronic Engineering, Xidian University, Xi’an, where he is currently the Director of Key Laboratory of Intelligent Perception and Image Understanding of the Ministry of Education of China. He has been a foreign member of the academia European and the Russian academy of natural sciences. His research interests include  machine learning, deep learning, natural computation, remote sensing, image processing, and intelligent information processing.
Prof. Jiao is the Chairman of the Awards and Recognition Committee, the Vice Board Chairperson of the Chinese Association of Artificial Intelligence, the fellow of IEEE/IET/CAAI/CIE/CCF/CAA, a Councilor of the Chinese Institute of Electronics, a committee member of the Chinese Committee of Neural Networks, and an expert of the Academic Degrees Committee of the State Council.
\end{IEEEbiography}

%\vspace{-1cm}
\begin{IEEEbiography}[{\includegraphics[width=1in,height=1.25in,clip,keepaspectratio]{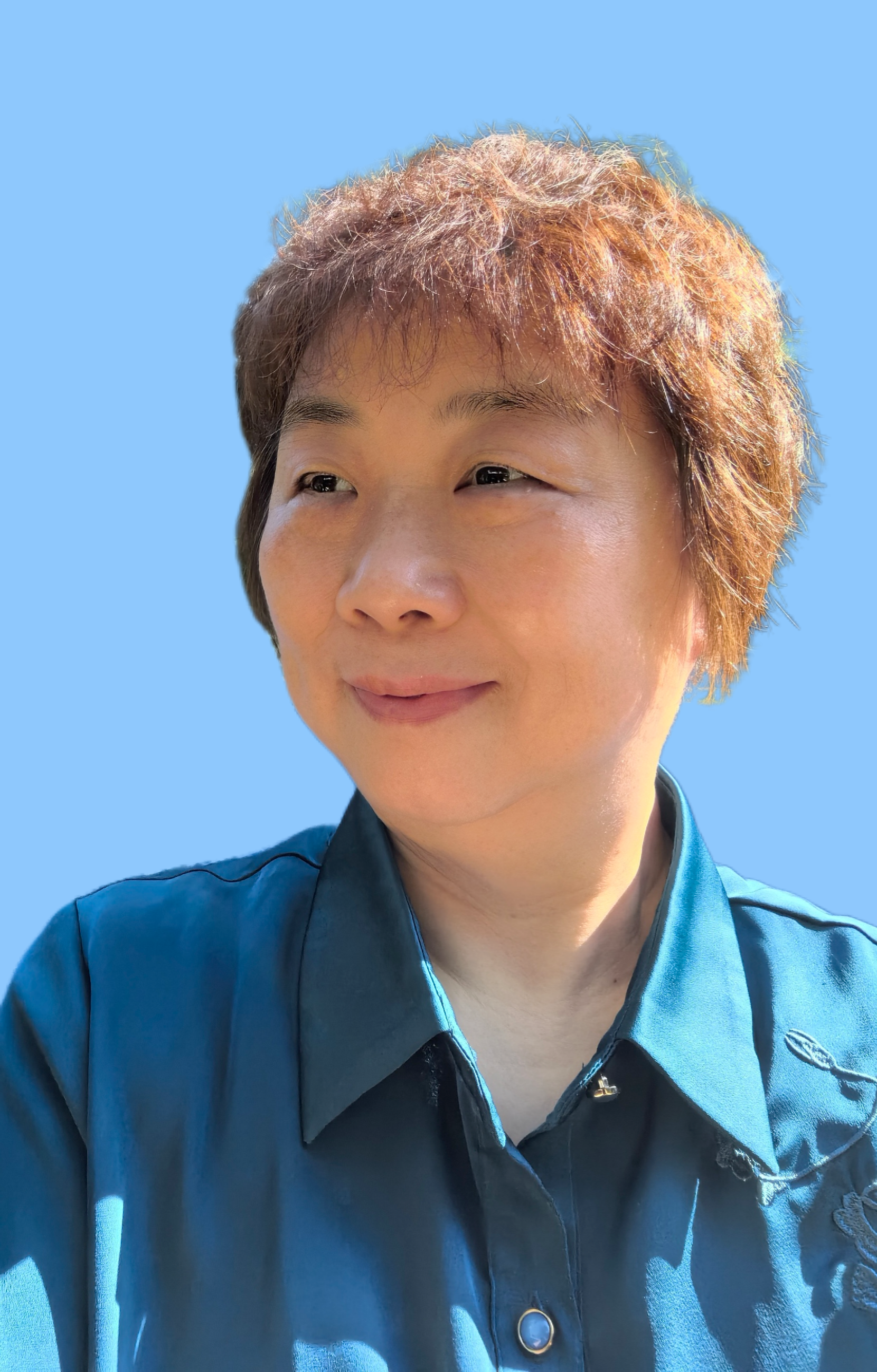}}]{Fang Liu} (Senior Member, IEEE) received a B.S. degree in computer science and technology from Xi’an Jiaotong University, Xi’an, China, in 1984 and the M.S. degree in computer science and technology from Xidian University, Xi’an, in 1995. She is currently a Professor at the School of Computer Science, Xidian University.
 Her research interests include signal and image processing, synthetic aperture radar image processing, multi-scale geometry analysis, learning theory and algorithms, optimization problems, and data mining.
\end{IEEEbiography}

%\vspace{-1cm}
\begin{IEEEbiography}[{\includegraphics[width=1in,height=1.25in,clip,keepaspectratio]{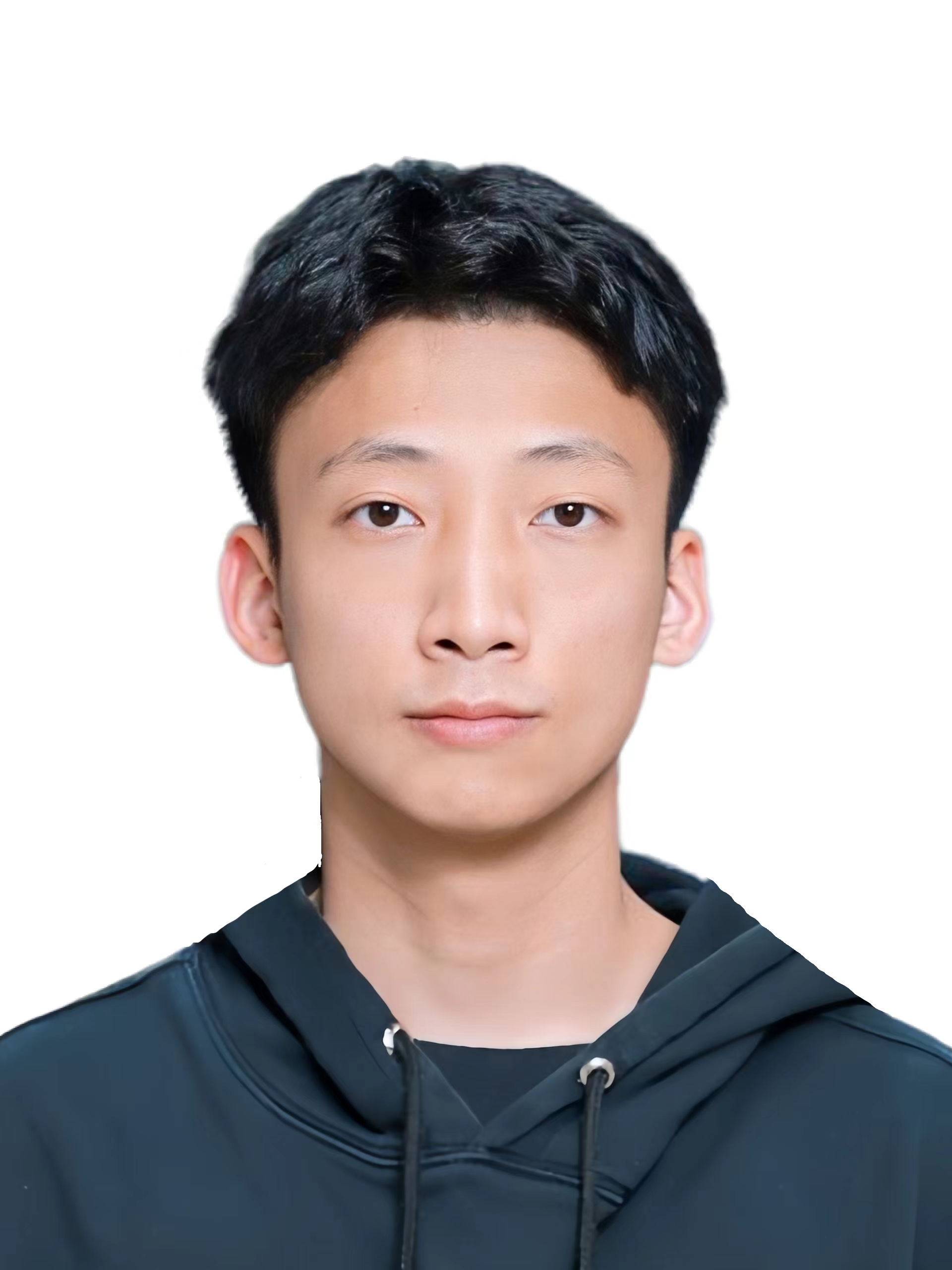}}]{Maoji Wen} entered the major of Computer Science and Technology at Xidian University in 2020, pursuing a Bachelor of Science degree. His research interests lie in computer vision.
\end{IEEEbiography}

%\vspace{-1cm}
\begin{IEEEbiography}[{\includegraphics[width=1in,height=1.25in,clip,keepaspectratio]{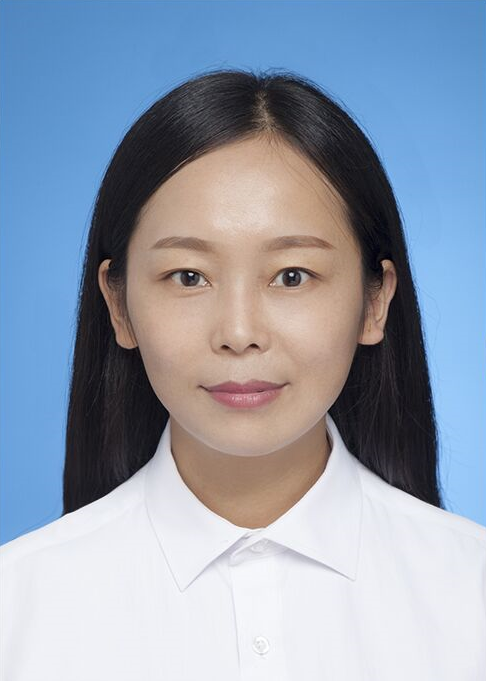}}]{Lingling Li} (Senior Member, IEEE) received the B.S. degree in electronic and information engineering and the Ph.D. degree in intelligent information processing from Xidian University, Xi’an, China, in 2011 and 2017, respectively. She is currently an Associate Professor with the School of Artificial Intelligence, Xidian University. From 2013 to 2014, she was an Exchange Ph.D. Student with the Intelligent Systems Group, Department of Computer Science and Artificial Intelligence, University of the Basque Country UPV/EHU, Leioa, Spain. Her research interests include quantum evolutionary optimization, and deep learning.
\end{IEEEbiography}

%\vspace{-1cm}
\begin{IEEEbiography}[{\includegraphics[width=1in,height=1.25in,clip,keepaspectratio]{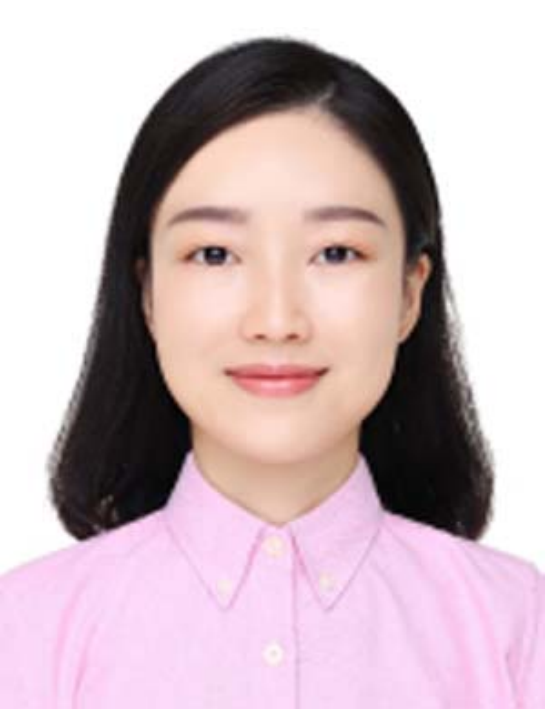}}]{Wenping Ma} (Senior Member, IEEE) received the B.S. degree in computer science and technology and the Ph.D. degree in pattern recognition and intelligent systems from Xidian University, Xi’an, China, in 2003 and 2008, respectively. She is currently an Associate Professor with the School of Artificial Intelligence, Xidian University. Her research interests include natural computing and intelligent image processing. Dr. Ma is a member of CIE.
\end{IEEEbiography}

%\vspace{-1cm}
\begin{IEEEbiography}[{\includegraphics[width=1in,height=1.25in,clip,keepaspectratio]{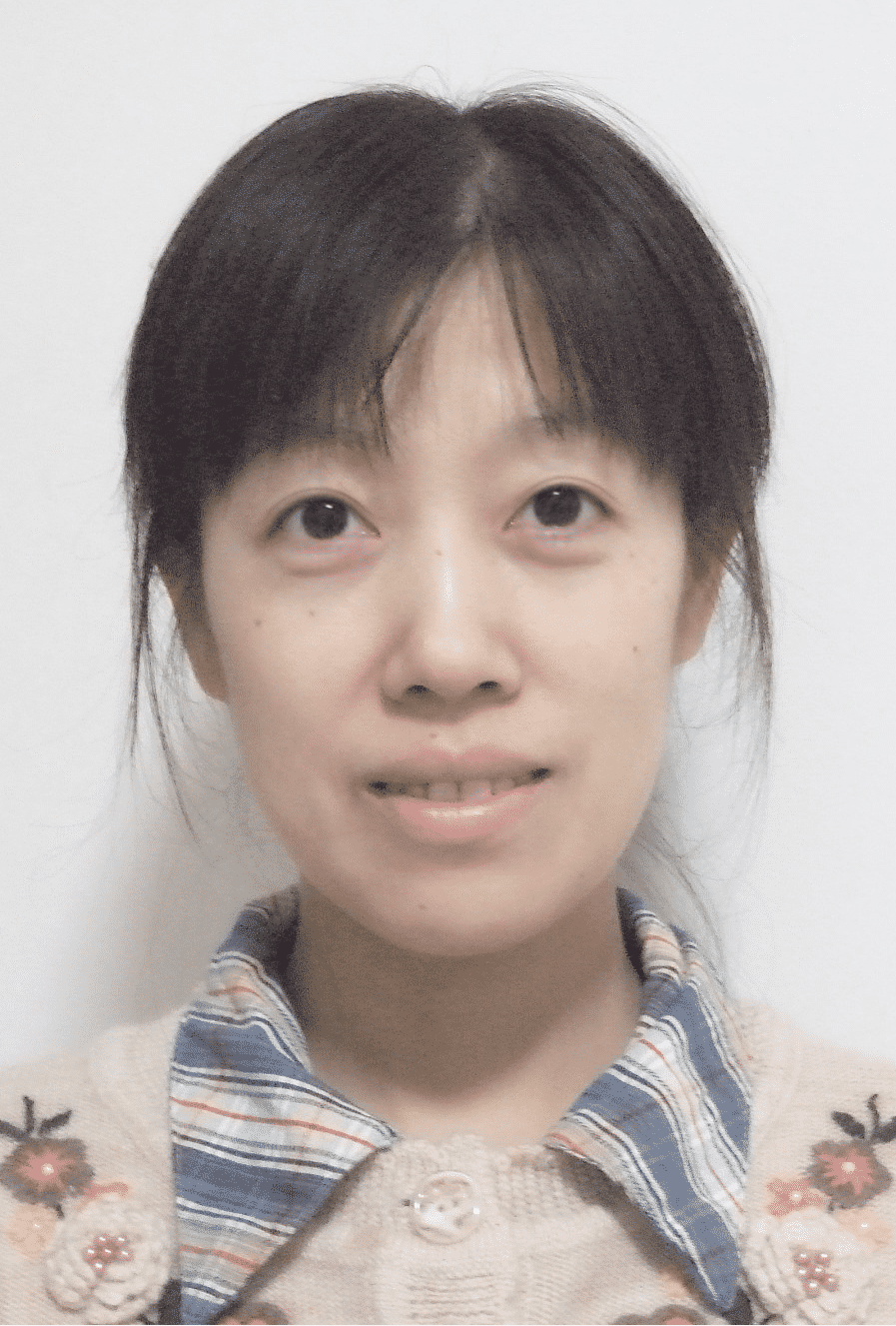}}]{Shuyuan Yang}
(Senior Member, IEEE) received the B.A. degree in electrical engineering and the M.S. and Ph.D. degrees in circuit and system from Xidian University, Xian, China, in 2000, 2003, and 2005, respectively. 
She has been a Professor with the School of Artificial Intelligence, Xidian University. Her research interests include machine learning and multiscale geometric analysis.
\end{IEEEbiography}

%\vspace{-1cm}
\begin{IEEEbiography}[{\includegraphics[width=1in,height=1.25in,clip,keepaspectratio]{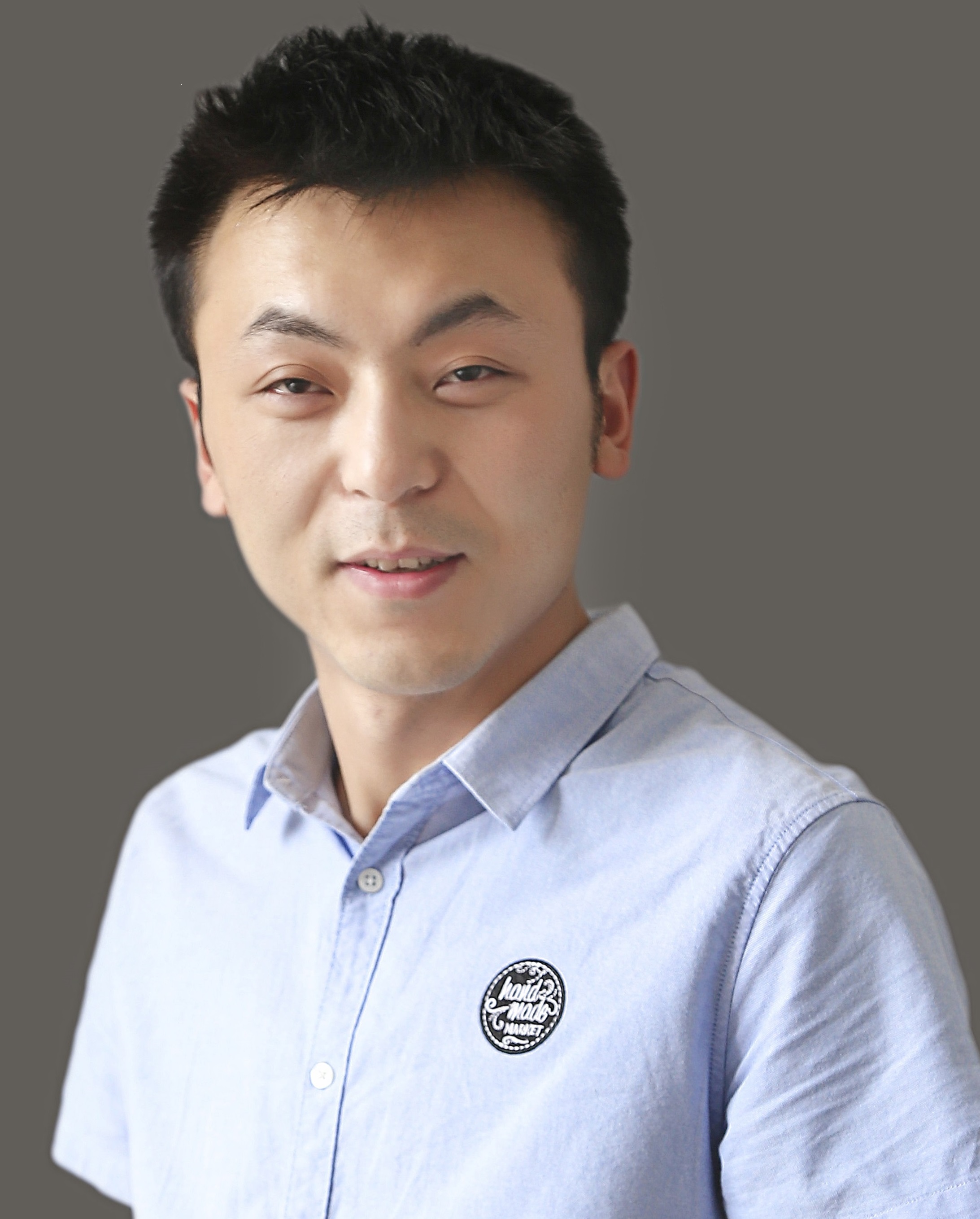}}]{Xu Liu}
Senior Member, IEEE) received the B.S. degree in mathematics and applied mathematics from the North University of China, Taiyuan, China, in 2013, and the Ph.D. degree from Xidian University, Xian, China, in 2019. 
He is currently an Associate Professor of Huashan elite and a Post-Doctoral Researcher with the Key Laboratory of Intelligent Perception and Image Understanding of Ministry of Education, School of Artificial Intelligence, Xidian University, Xi’an. He is the Chair of IEEE Xidian University Student Branch from 2015 to 2019. His research interests include machine learning and image processing.
\end{IEEEbiography}

%\vspace{-1cm}
\begin{IEEEbiography}[{\includegraphics[width=1in,height=1.25in,clip,keepaspectratio]{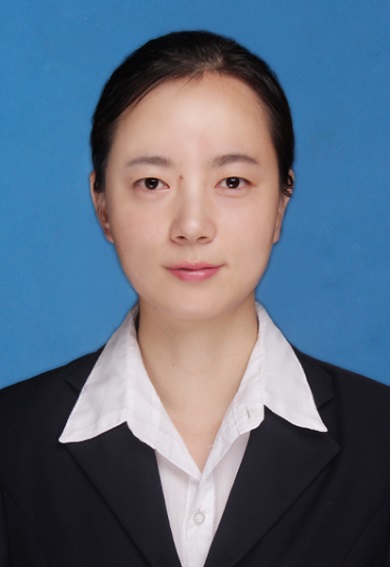}}]{Puhua Chen}
(Senior Member, IEEE) received the B.S. degree in environmental engineering from the University of Electronic Science and Technology of China, Chengdu, China, in 2009, and the Ph.D. degree in circuit and system from Xidian University, Xi’an, China, in 2016. She is currently an associate professor with the School of Artificial Intelligence, Xidian University. Her current research interests include machine learning, pattern recognition and remote sensing image interpretation.
\end{IEEEbiography}



%\bibliographystyle{IEEEtran}


%\vfill


\end{document}